\pdfoutput=1

\documentclass[11pt]{article}

\usepackage[]{ACL2023}

\usepackage{times}
\usepackage{latexsym}
\usepackage{booktabs, multirow} 
\usepackage{soul}
\usepackage{xspace} 
\usepackage{changepage,threeparttable} 
\usepackage{graphicx}
\usepackage{booktabs}
\usepackage{subfig}
\usepackage{tabularx}
\usepackage{enumitem}
\usepackage[T1]{fontenc}

\usepackage[utf8]{inputenc}

\usepackage{microtype}
\usepackage{amsfonts}

\usepackage{inconsolata}
\newcommand{\datasetname}{\textsc{InFoBench}\xspace}
\newcommand{\metricname}{\textsc{DRFR}\xspace}

\usepackage{booktabs, multirow} 
\usepackage{soul}
\usepackage{longtable}
\newenvironment{example}
  {\begin{quote}\itshape}
  {\end{quote}}
  
\title{\textsc{InFoBench}: Evaluating Instruction Following Ability \\in Large Language Models}

\newcounter{tencent}
\newcounter{ucf}
\newcounter{emory}
\newcounter{ug}
\newcounter{sjtu}
\author{
    Yiwei Qin$^{\Roman{tencent}}$, Kaiqiang Song$^{\Roman{tencent}}$, Yebowen Hu$^{\Roman{tencent},\Roman{ucf}}$, Wenlin Yao$^{\Roman{tencent}}$, Sangwoo Cho$^{\Roman{tencent}}$,
    \AND Xiaoyang Wang$^{\Roman{tencent}}$, Xuansheng Wu$^{\Roman{tencent},\Roman{ug}}$, Fei Liu$^{\Roman{emory}}$, Pengfei Liu$^{\Roman{sjtu}}$, Dong Yu$^{\Roman{tencent}}$ \\
    $^{\Roman{tencent}}$ Tencent AI Lab, Seattle; $^{\Roman{ucf}}$ University of Central Florida; \\
    $^{\Roman{emory}}$ Emory University; $^{\Roman{ug}}$ University of Georgia; $^{\Roman{sjtu}}$ Shanghai Jiao Tong University \\
    \small \texttt{qinyiwei07@outlook.com},
    \small \texttt{\{riversong, wenlinyao, swcho, shawnxywang, dyu\}@global.tencent.com} \\
    \small \texttt{yebowen.hu@ucf.edu},
    \small \texttt{xw54582@uga.edu},
    \small \texttt{fei.liu@emory.edu},
    \small \texttt{pengfei@sjtu.edu.cn}
}

\begin{document}
\maketitle

\renewcommand{\thefootnote}{\fnsymbol{footnote}}
\footnotetext[1]{Work was done during the internship of Yiwei Qin, Yebowen Hu and Xuansheng Wu at Tencent AI Lab.}
\renewcommand{\thefootnote}{\arabic{footnote}}

\begin{abstract}
This paper introduces the Decomposed Requirements Following Ratio (\metricname), a new metric for evaluating Large Language Models' (LLMs) ability to follow instructions. Addressing a gap in current methodologies, \metricname breaks down complex instructions into simpler criteria, facilitating a detailed analysis of LLMs' compliance with various aspects of tasks. Alongside this metric, we present \datasetname, a benchmark comprising 500 diverse instructions and 2,250 decomposed questions across multiple constraint categories. Our experiments compare \metricname with traditional scoring methods and explore annotation sources, including human experts, crowd-sourced workers, and GPT-4. The findings demonstrate \metricname's higher reliability and the effectiveness of using GPT-4 as a cost-efficient annotator. The evaluation of several advanced LLMs using this framework reveals their strengths and areas needing improvement, particularly in complex instruction-following. This study contributes a novel metric and benchmark, offering insights for future LLM development and evaluation.
\footnote{We release our code and dataset at \url{https://github.com/qinyiwei/InfoBench}.}
\end{abstract}

\section{Introduction}
\label{sec: intro}
The rapidly evolving domain of Natural Language Processing (NLP) has witnessed the advent and rise of Large Language Models (LLMs) \citep{ouyang2022training,brown2020language,openai2023gpt4,touvron2023llama,bai2022training,chowdhery2022palm,zhang2022opt}, which exhibit remarkable prowess in simulating human-like text generation.  Their ability to follow instructions is of utmost importance for their practical application, as it customizes the outputs according to specific user needs. Despite this, evaluation of these models frequently emphasizes general natural language tasks \citep{wang2018glue,wang2019superglue,hendrycks2020measuring,srivastava2022beyond,liang2022holistic}, or specific downstream tasks \citep{qin2023chatgpt,bang2023multitask}, while the critical element of instruction-following—the model's capacity to accurately understand and execute user instructions—has not been thoroughly explored. This evident gap in research and the absence of systematic evaluation methods dedicated to this crucial aspect serve as the impetus for our present study. We aim to establish a reliable protocol and benchmark for appraising the instruction-following aptitude of LLMs.

Current research tends to utilize evaluation methodologies such as A/B Testing \citep{askell2021general,alpaca,vicuna2023,alpaca_eval,wang2023pandalm}, 
overall scoring \citep{wang2022self}, and Elo rating \citep{zheng2023judging,bai2022training,glaese2022improving}—a derivative of A/B testing. While these strategies have proven somewhat effective, they suffer from considerable drawbacks, including scalability issues and a lack of interpretability. A/B testing, for instance, necessitates $N^2$ comparisons for N systems.
Moreover, the overall scoring system is opaque and difficult to interpret, failing to elucidate the reasoning behind the assignment of specific scores to particular instructions.

In this study, we propose a novel metric for assessing the instruction-following capability of LLMs, namely the Decomposed Requirements Following Ratio (\metricname).
Unlike traditional metrics, \metricname decomposes each instruction into simpler, distinct criteria, allowing for a more detailed and precise assessment of an LLM's ability to follow specific aspects of a given task. 
This decomposition enables a granular evaluation, where the compliance of the LLM is measured against each individual requirement within an instruction, offering a clearer picture of the model's performance in complex instruction-following scenarios.
In addition, We constructed a benchmark composed of 500 distinct instructions and provided 2,250 decomposed questions for the assessment of instruction-following. 
To facilitate a more fine-grained analysis, we classified the constraints present in each instruction into five categories: Content, Linguistic, Style, Format, and Number.
We also manually curated constraint labels for each decomposed question to identify and analyze potential areas of deficiency in each model. 

We conducted two key experiments: firstly, we compared the \metricname metric with traditional Direct Scoring (DS) for evaluating responses from various Large Language Models (LLMs).
The results showed a higher consensus among annotators with \metricname, particularly in \textit{Hard Set}, suggesting its greater reliability over Direct Scoring (\textbf{DS}).
Secondly, we explored more cost-efficient annotation sources by comparing human experts, crowd-sourced workers from Amazon Mechanical Turk (AMT), and GPT-4.
The findings revealed that while human experts were the most accurate, they were also costly and time-consuming.
AMT workers presented a balance but with lower accuracy.
Significantly, GPT-4 emerged as a highly accurate, cost-effective, and time-efficient alternative, especially when employing a structured, multi-turn prompt approach, making it a viable option for large-scale annotations.

We thus employ GPT-4 as an annotator to evaluate the entire \datasetname.
The comprehensive evaluation of these advanced LLMs reveals that while progress has been significant, there is still a notable gap in their ability to follow instructions perfectly, particularly in more complex scenarios represented by the "\textit{Hard Set}".
The closed-source models are currently leading, possibly due to better data or more sophisticated algorithms.
However, the performance across different constraint types and domains suggests that the challenges in instruction-following are more nuanced and may require focused improvements in areas like numerical and linguistic understanding.
As LLMs continue to evolve, addressing these specific areas could lead to more robust and versatile Large Language models capable of handling a broader range of real-world tasks and instructions.

Our study presents four key contributions to the field of evaluating Large Language Models (LLM) in their ability to follow instructions.
Firstly, we introduce a novel and scalable metric, the Decomposed Requirements Following Ratio (\metricname), which provides a reliable, detailed, and interpretable framework for assessing LLMs.
This metric decomposes complex instructions into simpler, specific criteria, allowing for a granular analysis of a model's performance.
Secondly, we have developed \datasetname, a comprehensive benchmark dataset.
It includes 500 diverse instructions paired with 2,250 decomposed questions, designed to test and analyze the instruction-following capabilities of LLMs systematically.
Thirdly, through extensive experimentation, we established the efficacy of \metricname and \datasetname.
Our findings highlight the potential of using GPT-4 as an evaluator for its accuracy and cost-effectiveness, particularly when employing structured, multi-turn prompts.
Finally, we utilized our automatic evaluation toolkit to conduct a thorough analysis of six advanced LLMs.
This evaluation not only underscores the current capabilities of these models but also sheds light on specific areas that require improvement for more nuanced and effective instruction-following, especially in complex scenarios.
These contributions pave the way for future advancements in LLM development and evaluation.
\section{\datasetname}
\label{sec: infobench}
Both evaluation metrics and benchmark datasets play an integral role in the objective assessment of Large Language Models' (LLM) ability to follow instructions.
These tools are essential for enabling consistent comparisons, pinpointing areas for improvement, and ensuring the practicality, fairness, and effectiveness of these models in real-world applications.
In Section~\ref{ssec: metric}, we introduce a novel evaluation metric, termed the \textbf{D}ecomposed \textbf{R}equirements \textbf{F}ollowing \textbf{R}atio (\metricname).
This metric is specifically designed to measure the proficiency of LLMs in adhering to complex instructions in a detailed and structured manner.
Furthermore, Section~\ref{ssec: dataset} discusses the development of a corresponding benchmark dataset, \datasetname, which has been meticulously curated to rigorously test and evaluate the instruction-following capabilities of LLMs.
This dataset presents a diverse range of scenarios and challenges, offering a comprehensive tool for benchmarking and refining the performance of these models.

\subsection{\metricname}
\label{ssec: metric}
\begin{table*}[htbp]
    \small
    \begin{tabular}{  p{4.5 cm} p{10.5cm}}
        \toprule
        \textbf{Instruction} & \textbf{Decomposed YES/NO Questions} \\
        \midrule
        Make a questionnaire to help hotel guests write hotel reviews.  
        & 1. Is the generated text a questionnaire?\newline
        2. Is the generated questionnaire designed for hotel guests?\newline
        3. Is the generated questionnaire helpful for hotel guests to write hotel reviews?
        \\\midrule
        Please generate 10 one-sentence hotel reviews from ten different customers, ensuring that 5 of the reviews are positive and 5 are negative. Begin each review with "CUSTOMER" and the customer's number. After completing the 10 reviews, provide a two-sentence summarization that captures the overall sentiment and key points from the reviews.    
        & 1. Does the generated text include hotel reviews?\newline
        2. Does the generated text include exactly 10  hotel reviews from 10 different customers?\newline
        3. Is each of the generated hotel reviews just one sentence long?\newline
        4. Are 5 of the reviews in the generated text positive and 5 negative?\newline
        5. Does each review in the generated text begin with the prefix "CUSTOMER" followed by the customer's number?\newline
        6. Does the generated text include a summarization after completing the 10 reviews?\newline
        7. Is the summarization in the generated text composed of two sentences?\newline
        8. Does the summarization in the generated text capture the overall sentiment and key points from the reviews?
        \\\bottomrule
    \end{tabular}
    \caption{Representative examples from the \datasetname. The first row illustrates an instance from the \textit{Easy Set}, while the second row presents a sample from the \textit{Hard Set}.}
    \label{tab: dataset examples}
\end{table*}

To evaluate the ability of advanced language models to accurately follow instructions, one common method is to consult evaluators for insights into whether and how the models' responses align with the given directives.
However, this approach often encounters difficulties in practice, as reaching a consensus among evaluators can be challenging, especially in scenarios where the models partially meet some aspects of the instructions.
A viable strategy for evaluating the compliance of advanced language models with complex instructions is to segment each instruction into distinct, simpler criteria that are more straightforward for evaluators to judge.
Consider a benchmark dataset comprised of $N$ instructions.
Each of these instructions can be decomposed into $m_i \in \mathbb{N}^+$ unique requirements.
The variable $r_{i,j} \in [0, 1]$ denotes whether the $j$-th criterion within the $i$-th instruction is met.
In this context, we can define the Decomposed Requirements Following Ratio (\metricname) as follows:
\begin{equation}
    \metricname = \frac{\sum_{i, j} r_{i,j}}{\sum_{i} m_i}
\end{equation}
This metric quantifies the model's adherence to the decomposed criteria across the dataset, offering a nuanced understanding of its instruction-following capabilities.
Table~\ref{tab: dataset examples} provides two illustrative examples of instructions, each accompanied by their respective decomposed criteria.
These criteria are framed as binary questions, necessitating a \textit{YES} or \textit{NO} response.
A \textit{YES} answer indicates that the output of the Large Language Model (LLM) successfully meets the specific criterion outlined in the question, corresponding to a value of $1$ for $r_{i,j}$.
It is important to highlight that even minor deviations from the requirement result in a negative (\textit{NO}) response, as precise adherence is mandatory.
For example, in the third question of the second example in Table~\ref{tab: dataset examples}, the occurrence of even a single hotel review exceeding one sentence in length will necessitate a \textit{NO} response.
On the other hand, a \textit{NO} indicates that the LLM’s output either does not satisfy the requirement or fails to provide relevant information pertinent to the specified criterion.

\subsection{\datasetname Dataset}
\label{ssec: dataset}
To effectively benchmark instruction-following abilities, it's crucial to develop a dataset with a broad spectrum of diverse instructions. This approach ensures thorough coverage and enables a more precise evaluation of proficiency in following instructions.
The \datasetname addresses this need with two distinct sets: the \textit{Easy Set} and the \textit{Hard Set}.
The \textit{Easy Set}, based on the research in \citep{wang2022self}, is designed for a broad range of applications.
In contrast, the \textit{Hard Set} is our original contribution, consisting of a manually curated dataset inspired by various subject areas.
Table~\ref{tab: dataset examples} showcases examples from both sets in the context of hospital reviews.
While the \textit{Easy Set} has been widely used in evaluations (\citep{wang2022self,alpaca_eval,alpaca,peng2023instruction,anand2023gpt4all,wang2023pandalm}), our work led to the creation of the \textit{Hard Set}, addressing the \textit{Easy Set}'s limitations:
\textbf{1)} Its instructions are relatively simple, limiting the ability to differentiate between modern LLMs' performance;
\textbf{2)} It features a limited variety of constraint types, typically with only about three per instruction, lacking in overall diversity;
\textbf{3)} The \textit{Easy Set} often includes ambiguous and subjective requirements, as exemplified by the difficulty in universally defining terms like ‘helpful’ in its instructions.
For instance, in Table~\ref{tab: dataset examples}, an instruction from the \textit{Easy Set} requires the questionnaire to be `helpful', but establishing a consensus on the definition of `helpful' proves to be difficult.

To guarantee the quality of the instructions and the clarity of the requirements, each instruction and its decomposed requirements are meticulously crafted by a subject matter expert and subsequently validated by another expert.
The process of collecting instances continues until both experts concur on the naturalness of the instruction and the atomicity and lucidity of the decomposed requirements.

\begin{figure}[htbp]
\centering
\includegraphics[width=7.5cm]{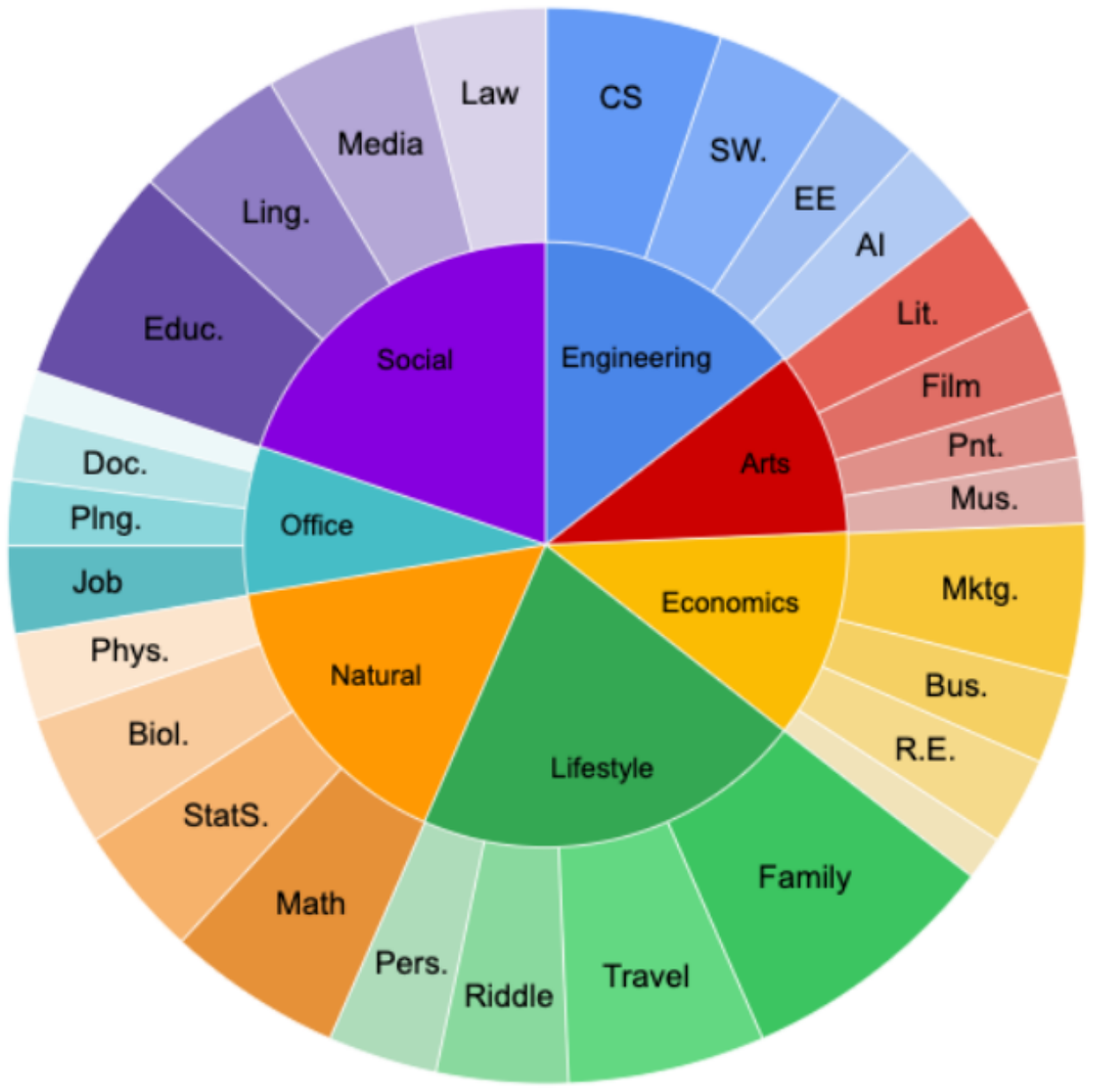}
\caption{Instruction category of the \textit{Hard Set}. Categories include: Engineering and Technology (Engineering), Arts (Arts), Business and Economics (Economics), Lifestyle and Leisure (Lifestyle), Natural Sciences (Natural), Office and Work (Office), Social Sciences (Social), Computer Science (CS), Software Engineering (SW.), Electrical Engineering (EE), Artificial Intelligence (AI), Literature (Lit.), Painting (Pnt.), Music (Mus.), Marketing (Mktg.), Business Administration (Bus.), Real Estate (R.E.), Human Resources (HR), Family and Relationships (Family), Travel and Tourism (Travel), Riddle and Puzzle (Riddle), Personal Development (Pers.), Mathmatics (Math), Statistics (StatS.), Biology (Biol.), Physics (Phys.), Job Recruitment (Job), Event Planning (Plng.), Document Formatting (Doc.), Calendar Management (Agd.), Education and Pedagogy (Educ.), Linguistics (Ling.), News and Media (Media), Law and Criminology (Law).
}
\label{fig: instruction_category}
\end{figure}

\subsubsection{Instruction Development}
\label{sssec: instruction development}
Our objective is to formulate instructions with extensive coverage that effectively distinguish between superior and inferior models.
We have developed the \textit{Hard Set} encompassing 72 domains, spanning areas such as Natural Sciences, Social Sciences, Engineering, Economics, Arts, various occupational tasks, and aspects of daily life, illustrated in Figure~\ref{fig: instruction_category}.

We also incorporate specific response constraints in these instructions:
1) \textbf{Content Constraints}:
These define the specific topics or details to be addressed in the LLM's response, such as discussing "\texttt{climate change impacts}" or mentioning "\texttt{the main characters in Harry Potter}";
2) \textbf{Linguistic Guidelines}:
These dictate the use of particular language structures and terms, including grammatical styles, syntax, and specific dialects, like "\texttt{Victorian English}" or "\texttt{technical jargon}";
3) \textbf{Style Rules}:
These direct the overall tone and audience of the text, varying from formal to persuasive or sophisticated, as in writing with a "\texttt{respectful tone}" or for "\texttt{a young audience}";
4) \textbf{Format Specifications}: These instruct the LLM on the structural presentation of its response, such as "\texttt{crafting a sonnet}" or "\texttt{list ideas bullet-wise}";
5) \textbf{Number Limitations}:
These involve numeric-related instructions, like producing "\texttt{a 500-word essay}" or presenting "\texttt{three arguments for renewable energy}".

\subsubsection{Instruction Decomposition}
\label{sssec: instruction decomposition}
In the process of instruction development, we meticulously devised detailed requirements and associated binary questions for each instruction, each question reflecting a specific constraint that can be answered with "YES" or "NO".
Utilizing our established ontology of constraint types, we manually classified each question with appropriate constraint labels, noting that a single requirement could embody multiple constraint types.
For example, the question "\texttt{Does the generated text provide a critique of Vincent van Gogh's Starry Night?}" falls under the content constraint due to its focus on "\texttt{Starry Night}", while also fitting the format constraint as a critique represents a distinct writing style. Therefore, we assigned one or more constraint types to each question, with each constraint in our dataset being labeled with up to three types.

\subsubsection{Benchmark Statistics}
\label{sssec: benchmark statistics}
\begin{table}[htbp]
    \centering
    \small
    \begin{tabular}{lrrrrrr}
        \toprule
        & \textbf{\#Inst.} & \textbf{Len.} & \textbf{\#Req.} & \textbf{\#R/I} & \textbf{\#Dom.} \\\midrule
        Easy &252 &17.77 &690 &2.74 &71 \\
        Hard &248 &59.26 &1560 &6.29 &72 \\\midrule
        Overall &500 &38.35 &2250 &4.50 &143 \\
        \bottomrule
    \end{tabular}
    \caption{Statistics of \datasetname Dataset: This table provides key statistics including the number of instructions (\textbf{\#Inst.}), the average length of instructions (\textbf{Len.}), the total number of decomposed questions in the dataset (\textbf{\#Req.}), number of requirements per instruction (\textbf{\#R/I}), and the number of instruction domain (\textbf{\#Dom.}).
    }
    \label{tab: statistics}
\end{table}
Table~\ref{tab: statistics} offers statistical insights into the \datasetname, highlighting significant differences between the two sets.
The \textit{Hard Set} is characterized by more elaborate instructions and a greater number of requirements, indicating a higher level of complexity compared to the \textit{Easy Set}.
Moreover, the table emphasizes the dataset's diversity, evident in the broad range of domains it encompasses.

\begin{figure}[htbp]
\centering
\includegraphics[width=7.5cm]{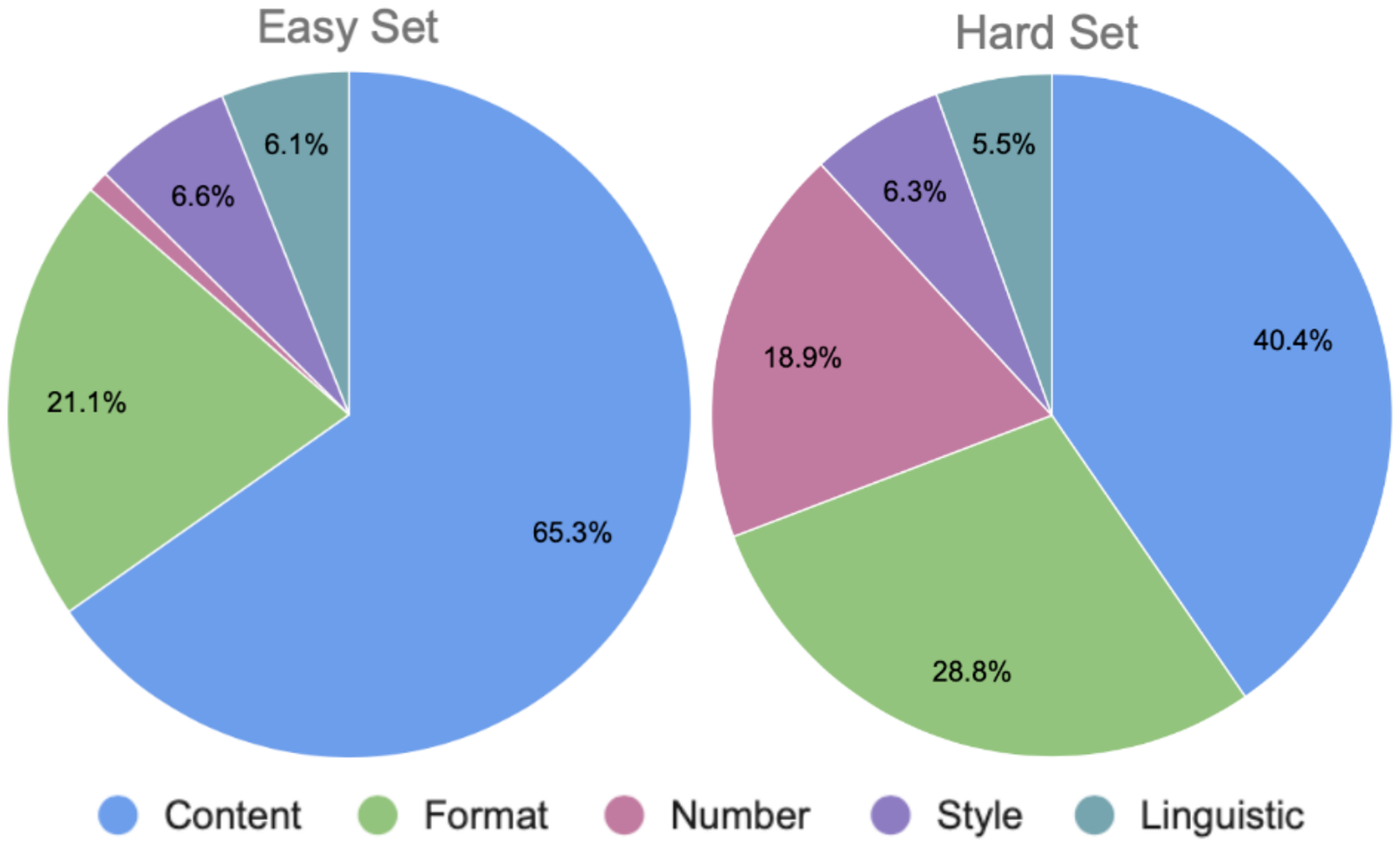}
\caption{Constraint types of the Easy and the Hard Set.
} 
\label{fig: constrain_type}
\end{figure}
Figure \ref{fig: constrain_type} illustrates a comparative analysis of the constraint type distribution between the \textit{Easy Set} and the \textit{Hard Set}. Approximately two-thirds of the constraints in the \textit{Easy Set} are classified as content constraints. In contrast, the \textit{Hard Set} demonstrates a more heterogeneous distribution of varied constraint types.
More detailed statistics are elaborated in Appendix~\ref{sec: appendix_instruction_dataset_statistics}.

\section{Experiments}
\label{sec: experiments}
\subsection{\metricname v.s. Direct Scoring}
\label{ssec: exp on drfr}
In our study, we aim to assess the reliability of the \metricname metric by contrasting it with a conventional direct scoring approach.
In the Direct Scoring (\textbf{DS}), human evaluators assign a score ranging from 1 (indicating the lowest quality) to 5 (representing the highest quality) for each response generated by the model.
The criteria for each rating are elaborated as follows:
\textbf{Rating-1 (Very Bad)}:
The model's response is entirely irrelevant to the query posed. It demonstrates no understanding of the requested task.
\textbf{Rating-2 (Bad)}:
While there is a minimal degree of relevance to the original instruction, the response is predominantly incorrect, confusing, or both.
\textbf{Rating-3 (Okay)}:
The response shows some relation to the posed question but lacks critical details or includes inaccuracies.
\textbf{Rating-4 (Good)}:
This rating is given when the response is mostly accurate and comprehensive. However, it might include slight errors or omit minor details.
\textbf{Rating-5 (Excellent)}:
A response receiving this rating is entirely accurate, detailed, and aligns perfectly with the requirements of the query.

In our evaluation, we incorporate five widely recognized Large Language Models (LLM), each distinguished by its unique characteristics and training methodologies:
\noindent \textbf{GPT-3.5-turbo}~\citep{ouyang2022training}:
A 175 billion parameter extension of GPT-3~\citep{brown2020language} by OpenAI, enhanced for alignment with user intentions through human feedback-based reinforcement learning.
\noindent \textbf{GPT-4}~\citep{openai2023gpt4}:
This model is an advanced, large-scale successor to GPT-3.5-turbo.
\noindent \textbf{Claude-v1}~\citep{bai2022training}:
Anthropic's 52 billion parameter AI assistant, Claude-v1, is trained to be helpful, honest, and harmless using reinforced learning from human feedback.
\noindent \textbf{Alpaca-7B}~\citep{alpaca}:
Alpaca-7B is an open-source model, derived from a 7 billion parameter LLaMA base~\citep{touvron2023llama}, and fine-tuned with a dataset~\citep{wang2022self} focused on instruction-following capabilities.
\noindent \textbf{Vicuna-13B}~\citep{vicuna2023}:
Vicuna-13B is a 13 billion parameter, open-source model fine-tuned from LLaMA, incorporating diverse real-world interactions from the ShareGPT platform\footnote{\url{https://sharegpt.com}}.

Additionally, We aggregate responses generated by the models on \datasetname and engage three NLP experts to independently assess them, utilizing two distinct methods.
Initially, they apply the Direct Scoring (\textbf{DS}), followed by an evaluation using the \metricname metric.
This order is designed to mitigate any potential bias: by conducting the \metricname assessment after the DS evaluation, we aim to prevent the evaluators' impressions formed during the \metricname assessment from influencing their \textbf{DS} evaluations.
This ensures a more objective and unbiased evaluation process.

Moreover, to facilitate an equitable comparison between the two evaluation metrics, we standardize both into instruction-level scores.\footnote{\textbf{DS} uses the score provided by the evaluator, \metricname counts all the stratified requirements and normalized by the number of requirements in the instruction}
This enables a direct, side-by-side comparison of response pairs from two distinct models, referred to as Model-A and Model-B.
For each instructional prompt, the paired responses are categorized into three distinct classifications: 1) Model-A outperforms Model-B, 2) Model-A and Model-B are equally effective, and 3) Model-B outperforms Model-A.
Finally, a Fleiss' Kappa Agreement~\citep{fleiss1971measuring} is employed to measure the agreements among three evaluators concerning the pairwise categorization.
This \textbf{Pairwise Kappa Agreement} approach ensures clarity and precision in our comparative analysis of the evaluation metrics.

\begin{table}[hbtp]
    \centering
    \small
    \begin{tabular}{lrrr}
    \toprule
     & \multicolumn{3}{c}{Pairwise Kappa Agreement} \\
    Metric & \textit{Easy}(25) & \textit{Hard}(25) & Overall(50) \\
    \midrule
    DS & 0.241 & 0.302 & 0.284 \\
    \metricname & \textbf{0.537} & \textbf{0.493} & \textbf{0.532} \\
    \bottomrule
    \end{tabular}
    \caption{Pairwise Kappa Agreement of \textbf{DS} and \metricname on 50 instructions from \datasetname.
    A score below 0.2 suggests a slight agreement
    A score of 0.21 to 0.40 suggests a fair agreement.
    A score of 0.41 to 0.60 suggests a moderate agreement.
    A score of 0.61 to 0.80 suggests a substantial agreement.
    And a score of 0.81 to 1.00 suggests an almost perfect agreement.
    }
    \label{tab: metric experiment}
\end{table}

Table~\ref{tab: metric experiment} displays the Pairwise Kappa Agreement for both metrics across 50 selected instances from the \datasetname, comprising an equal mix of 25 instances each from the Easy and Hard subsets.
The observed gap exceeding 0.19 between the \metricname and Direct Scoring (\textbf{DS}) suggests that \metricname yields a notably higher annotator consensus, thereby implying its enhanced reliability for human annotations.

\subsection{Experiment of Annotation Sources}
\label{ssec: exp annotation sources}
In our quest to identify a more cost-effective source for annotations, we conducted a comparative study utilizing annotations from various sources, including human experts, crowd-sourced workers from Amazon Mechanical Turk (AMT), and advanced Large Language Models like GPT-4. The details of this comparison are as follows:

\noindent \textbf{Human Expert Annotations}:
Annotations from three experts, previously discussed in Section~\ref{ssec: exp on drfr}, were utilized.
Given the wide range of topics covered by the instruction set, it is improbable for a single expert to possess comprehensive proficiency across all subjects.
Therefore, we implemented a majority voting system, supplemented by the use of online research tools, to enhance the accuracy of these expert annotations.

\noindent \textbf{Crowd-Sourced Annotations}:
We engaged annotators from the Amazon Mechanical Turk (AMT) platform\footnote{https://www.mturk.com/}, selecting those with a HIT Approval Rate above 90\% and a minimum of 500 Approved HITs.
Each task, comprising an instruction and its corresponding responses, was assigned to three distinct annotators.
We adopted a majority voting approach for each discrete query and an aggregate scoring method for instructions.
The specifics of this evaluation methodology are elaborated in Appendix \ref{sec: appendix_human_annotation_guidance}.

\noindent \textbf{GPT-4 Based Annotation}:
We employed the GPT-4 model\footnote{Version: gpt-4-0314.} for automatic evaluation.
The approach involved issuing a series of multi-turn prompts, sequentially presenting GPT-4 with decomposed questions, while accounting for the context provided by previous answers. 
The specifics of this prompt structure are detailed in Appendix~\ref{sec: appendix_prompt_design}.

\begin{table*}[htbp]
    \centering
    \small
    \setlength\tabcolsep{5pt}
    \begin{tabular}{llrrrrrrr}
        \toprule
        & & & & \multicolumn{5}{c}{Robustness} \\
        \cmidrule{5-9}
        & \textbf{Source} & \textbf{Cost}(\$) & \textbf{Time}(min) & \textbf{ACC}(\%) & \textbf{PLD0}(\%) & \textbf{PLD1}(\%) & \textbf{PLD2}(\%) & \textbf{WPLD} \\
        \midrule
        \multirow{3}{*}{\textit{Easy Set}(25)} & Expert & 472.00 & 480 & 100.0 & 100.0 & 0.0 & 0.0 & 0\\
        &AMT & 93.75 & 960 & 90.0 & 51.8 & 37.6 & 10.6 & 0.588 \\
        &GPT-4 & 5.40 & 20 & \textbf{93.0} & 65.6 & 34.4 & 0.0 & \textbf{0.344} \\
        \midrule
        \multirow{3}{*}{\textit{Hard Set}(25)} & Expert & 944.00 & 960 & 100.0 & 100.0 & 0.0 & 0.0 & 0 \\
        & AMT & 93.75 & 1,200 & 81.0 & 51.4 & 31.6 & 17.0 & 0.656\\
        &GPT-4 & 14.20 & 40  & \textbf{87.0} & 52.4 & 41.6 & 6.0 & \textbf{0.536} \\
        \midrule
        \multirow{3}{*}{Overall(50)} &Expert & 1,416.00 & 1,440 & 100.0 & 100.0 & 0.0 & 0.0 & 0\\
        &AMT & 187.50 & 2,160 & 84.0 & 51.6 & 34.7 & 13.7 & 0.621 \\
        &GPT-4 & 19.40 & 60 & \textbf{89.0} & 59.0 & 38.0 & 3.0 & \textbf{0.440} \\
        \bottomrule
    \end{tabular}
    \caption{
    Comparative Analysis of Annotation Sources for 50 Instructions in \datasetname across five models: Cost, Time, and Robustness Evaluation.
    \textbf{AMT} stands for Amazon Mechanical Turk.
    All the annotations are aggregated by a majority vote of three annotators except the GPT-4. 
    \textbf{ACC} compares the accuracy of the annotations against a benchmark 'ground truth' label, which is derived from an aggregated human expert label.
    \textbf{PLD}(Pairwise Label Distance) is a critical measure reflecting the level of agreement in annotations on an instruction level.
    \textbf{WPLD} is the weighted average of PLD for all instructions.
    }
    \label{tab: experiment_annotation}
\end{table*}
In Table~\ref{tab: experiment_annotation}, we present a comparative analysis of various annotation sources, emphasizing their cost, time efficiency, and annotation robustness. 
We calculated the cost of expert annotations by aggregating the total hours worked by three NLP experts.
The hourly rate was set at \$59, based on salary data from ZipRecruiter.\footnote{\url{https://www.ziprecruiter.com/Salaries/NLP-Salary}}
The expenditure on AMT workers was computed based on the total payment made on the platform. Each annotation task, comprising one instruction and five model responses, was assigned to three annotators.
The compensation was \$1 per annotator, with an additional 25\% platform fee.
The total time for AMT annotations was obtained from AMT platform statistics.
The cost of using the GPT-4-0314 model is determined by its token-based pricing structure\footnote{\url{https://platform.openai.com/docs/deprecations}}, which is \$0.03 per 1,000 prompt tokens and \$0.06 per 1,000 completion tokens.
Operational constraints of GPT-4 include a limit of 200 requests per minute and a maximum of 40,000 tokens per minute.\footnote{\url{https://platform.openai.com/docs/guides/rate-limits?context=tier-free}}
The total cost and time expenditure for using GPT-4 were calculated based on the cumulative count of prompt and completion tokens used during the execution of the tasks.
To evaluate the robustness of the annotation source, we inherit the pairwise comparison mentioned in Section~\ref{ssec: exp on drfr}.
The three categories are labeled with $\{-1, 0, 1\}$ for 1) Model-A outperforms Model-B, 2) Model-A and Model-B are equally effective, and 3) Model-B outperforms Model-A.
Pairwise Label Distance(\textbf{PLD}) is introduced to measure the agreement between the annotations and the ground truth labels.
A PLD value of 0 indicates that the annotations accurately reflect the relative performance order of the models for a given pair of responses.
A PLD of 1 suggests a potential misclassification in scenarios where models perform equally (‘tie’ scenario), possibly attributing superiority to one model over the other, or vice versa.
A PLD of 2 signifies that the annotations incorrectly represent the order of model performances.
A Weighted Pairwise Label Distance (\textbf{WPLD)} is defined as below:
\begin{equation}
    \mathbf{WPLD} = \sum_{i=0}^{2} i * \mathbf{P}(\mathbf{PLD} = i)
\end{equation}

Our analysis yields two important findings:
\textbf{1)} Annotations generated by GPT-4 exhibit superior cost-effectiveness and time efficiency compared to those produced by AMT (Amazon Mechanical Turk) crowd workers. Additionally, GPT-4 demonstrates enhanced overall performance in annotation tasks.
\textbf{2)} When employing decomposed questions, GPT-4's performance is particularly noteworthy. It achieves an accuracy rate of 89\% and a weighted pairwise label distance of 0.44, in alignment with the gold standard. These results suggest that GPT-4, used in conjunction with the decomposed questions, emerges as a feasible and efficient alternative to traditional expert-based annotation, especially when considering the balance between cost and time efficiency.
In light of these findings, we emphasize the use of the GPT-4 model for annotation tasks in our evaluation of models on \datasetname, considering its advantageous balance of cost, time efficiency, and accuracy.

\section{Automatic Evaluation}
\label{sec: automatic evaluation}
In light of the findings from our previous experiments, it has become evident that GPT-4 represents a superior alternative for expert annotations.
This approach not only offers a high degree of accuracy but also ensures time efficiency and cost-effectiveness in annotation processes. Building on this insight, we integrate the gpt-4-0314 with the multi-turn prompt for scaling up the evaluation across the entire \datasetname with six of the most cutting-edge LLMs.
These include: gpt-3.5-turbo-1106~\cite{brown2020language} and gpt4-1106-preview~\cite{gpt4-turbo} released by OpenAI, Claude-2.1~\cite{mixtral-8x7b} by Anthropic, Gemini-Pro~\cite{geminiteam2023gemini} by Google, Llama-2-70b-chat~\cite{touvron2023llama2} by Meta and Vicuna-13b-v1.5~\cite{zheng2023judging} by LMSYS.
To avoid randomness during the decoding process, we use greedy decoding\footnote{$temperature = 0, top\_p = 1$} for all the models.

\begin{table}[htbp]
    \centering\
    \small
    \setlength\tabcolsep{4.5pt}
    \begin{tabular}{l|ccc}
    \toprule
    Model Name & \textit{Easy Set} & \textit{Hard Set} & Overall \\
    \midrule
    gpt-4-1106-preview & 90.1 & 89.1 & 89.4 \\
    gpt-3.5-turbo-1106 & 90.4 & 85.1 & 86.7 \\
    claude-2.1 & 82.9 & 87.9 & 86.4 \\
    gemini-pro & 87.3 & 84.9 & 85.6 \\
    Llama-2-70b-chat & 89.6 & 82.1 & 84.4 \\
    Vicuna-13b-v1.5 & 86.1 & 75.0 & 78.4 \\
    \bottomrule
    \end{tabular}
    \caption{Automated \metricname assigned by gpt-4-0314 for six selected LLMs on the \datasetname.}
    \label{tab: gpt4_evaluation_result}
\end{table}

\noindent \textbf{Overall}:
Table~\ref{tab: gpt4_evaluation_result} presents the automated \metricname scores powered by gpt-4-0314 for the selected models on \datasetname.
Upon analysis of the data presented in the table, three key observations can be made:
(1) GPT-4 exhibits superior performance over the other LLMs, scoring the highest amongst all.
However, even with this state-of-the-art model, more than 10\% of the requirements remain unfulfilled, indicating a critical need for further enhancement within the aspect of instruction following.
(2) All models except Claude-2.1 display better performance on the \textit{Easy Set} as compared to the \textit{Hard Set}, implying that the \textit{Hard Set} poses a more formidable challenge. 
(3) The closed-sourced models (e.g. gpt-4-1106-preview, gpt3.5-turbo-1106, claude-2.1, gemini-pro) have better performance on \textit{Hard Set} than those open-sourced models (e.g. Llama-2-70b-chat, Vicuna-13b-v1.5), which may indicate they have better data instruction coverage on more advanced instructions or they have better model generalization capabilities.

\begin{figure}[htbp]
    \centering
    \includegraphics[width=\linewidth]{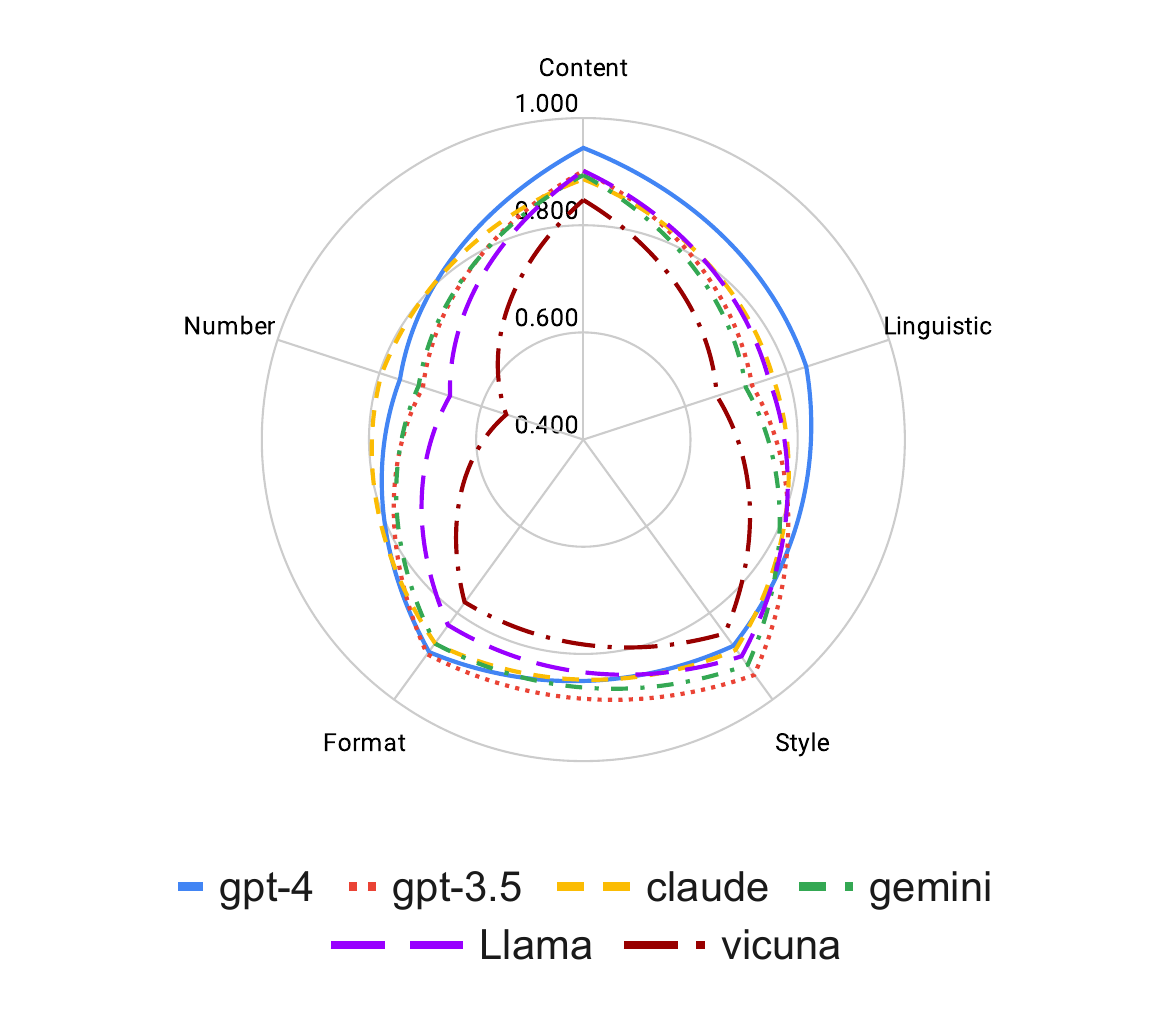}
    \caption{Radar plot comparing the performance of LLMs on different constraint types powered by gpt-4-0314}
    \label{fig: constraint type}
\end{figure}
\noindent \textbf{Constraint Type}:
To assess the influence of constraint type on adherence to instructions, we calculate the evaluation scores for each constraint type individually, as presented in Figure~\ref{fig: constraint type}. 
The radar plot reveals a discernible pattern in performance across different constraint types: performance is highest on Content and Style, intermediate on Format, and lowest on Number and Linguistic constraints.
This general trend is observed across all six language models.

\begin{figure}[htbp]
    \centering
    \includegraphics[width=\linewidth]{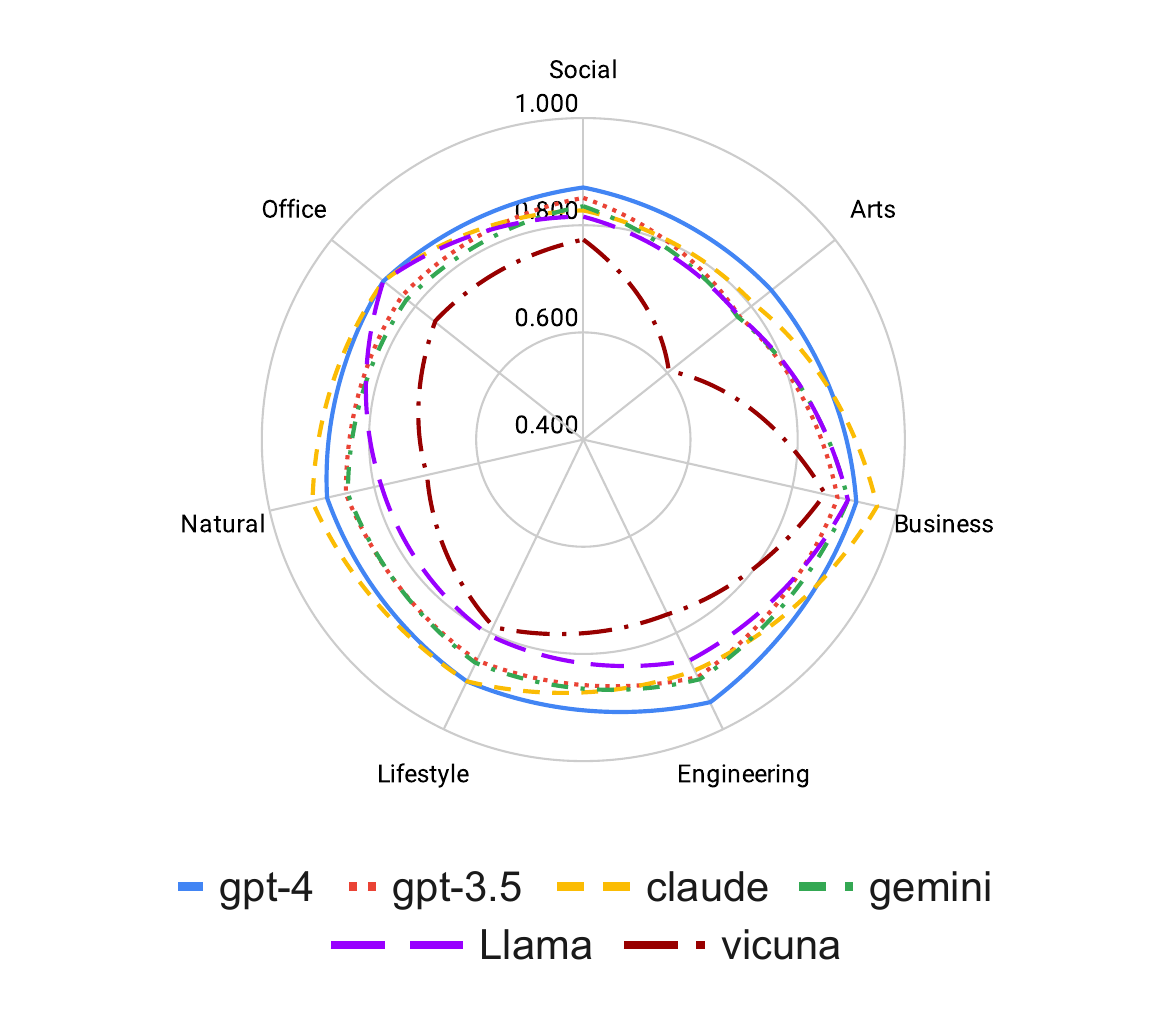}
    \caption{Radar plot comparing the performance of LLMs on different domains powered by gpt-4-0314}
    \label{fig: domain}
\end{figure}
\noindent \textbf{Domain}:
To investigate the influence of the instruction domains on the behavior of instruction-following, we restrict our analysis to the seven principal subject categories within the \textit{Hard Set}, and independently calculate the evaluation scores for each instruction category, as depicted in Figure~\ref{fig: domain}.
Contrary to the pattern discernible in the preceding section, the resulting shape of the evaluation scores on the radar plot approximates a circle (except for vicuna).
This indicates that the domain of instruction does not serve as the principal factor affecting the instruction-following or most of the models have a good data coverage for such domains.

\section{Related Work}
\label{sec: related work}
The adaptation of LLMs for enhanced usability has garnered significant attention. Two prevalent approaches for this adaptation are instruction tuning and alignment tuning \citep{zhao2023survey}. Instruction tuning \citep{wang2022self,lou2023prompt} aims to unlock the capabilities of LLMs by fine-tuning them on annotated “instructional” data. This involves datasets with natural language instructions and their corresponding desired outcomes. Several LLM adaptations have embraced this method for effective refinement\citep{alpaca,wei2021finetuned,chung2022scaling,iyer2022opt,ouyang2022training}. Alignment tuning \citep{ziegler2019fine,christiano2017deep,ouyang2022training,bai2022training} differs from instruction tuning by emphasizing human feedback to ensure LLMs align with human values and preferences. This approach addresses LLMs' tendencies to exhibit unintended behaviors like fabricating information or producing biased expressions\citep{ouyang2022training,kenton2021alignment}, steering them more in line with human expectations.

A natural extension of this line of inquiry is assessing LLMs’ ability to accurately follow natural language instructions and align with human preferences. In terms of evaluation, using LLMs for Natural Language Generation (NLG) evaluation has shown promise in numerous tasks \citep{zheng2023judging, liu2023geval, wang2023chatgpt, fu2023gptscore, kocmi2023large,xu2023wizardlm, vicuna2023,eldan2023tinystories}. Given this context, in our proposed work, we employ LLMs as evaluators to assess their aptitude for instruction-following.

Several works have concurrently assessed LLMs from various perspectives: 
\citet{wang2023scibench} explore the reasoning capabilities of LLMs for complex scientific problem solving;
\citet{wang2023pandalm} present a judge LLM trained to distinguish the performance of various LLMs;
\citet{li2023instructionfollowing} introduce an instruction-following evaluation protocol named verbalizer manipulation on classification tasks;
\citet{alpaca_eval} develop a LLM-based automatic evaluation method to calculate win-rate of the given model's output favored over that of a reference model on an 805-instruction dataset \citep{dubois2023alpacafarm};
\citet{xu2023wizardlm} propose an automatic method to compose complex instructions for instruction-tuning;
\citet{chia2023instructeval} introduce an evaluation suite designed for the assessment of instruction-tuned LLMs' ability on problem-solving, writing proficiency, and alignment with human values.

Two concurrent studies align with our efforts in establishing benchmarks for evaluating LLMs' instruction-following abilities. In contrast to our method of decomposing a single instruction into multiple constraints, \citet{jiang2023followbench} adopts a distinct approach by sequentially adding fine-grained constraints to construct multi-level instructions. Meanwhile, \citet{zhou2023instructionfollowing} emphasizes objective evaluations with 25 verifiable instructions, contrasting with our work's broader coverage yet similar focus on objectivity.
\section{Conclusion}
\label{sec: conclusion}

This study introduces a significant breakthrough in evaluating Large Language Models (LLMs) through the Decomposed Requirements Following Ratio (\metricname), a novel metric that dissects complex instructions into simpler criteria for a more nuanced analysis. This approach, coupled with the creation of \datasetname, a comprehensive benchmark dataset, enables a detailed assessment of LLMs' instruction-following capabilities. Our findings reveal the superiority of \metricname over traditional evaluation methods, offering clearer insights into model performance, especially in complex scenarios. The use of GPT-4 as an annotator emerged as a cost-effective, accurate, and efficient alternative for large-scale evaluations, highlighting its potential in automating and enhancing the assessment process.

The comprehensive evaluation of various LLMs using our methods uncovered their strengths and areas needing improvement. While significant progress is evident, especially in closed-source models, challenges remain in handling complex instructions, signaling a need for advancements in areas like numerical reasoning and linguistic comprehension. Our contributions provide a foundation for future research and development, guiding the enhancement of LLMs for more accurate and reliable performance in real-world applications. This study not only underscores the current capabilities of LLMs but also paves the way for the next generation of more robust and versatile AI models.

\section*{Limitations}
\label{sec: limitations}
This study is not without its limitations, which must be acknowledged:

1. The cost and resources required for human annotation have restricted us to providing these annotations for only a fraction of our instruction set. Specifically, a mere 10\% of the instruction set, equating to 50 instructions, were annotated by humans. The reliability of our decomposed protocol, comparisons with the baseline protocol, and comparisons among different annotations are, therefore, based solely on this subset. 

2. The size of our dataset presents another limitation. It comprises 500 instructions and 2,250 decomposed questions, representing a somewhat limited benchmark. The manual nature of our instruction writing process limits our capacity to scale this dataset significantly. Future work could explore the development of automated methods for instruction writing and question decomposition, potentially allowing for the creation of more scalable datasets.

3. Finally, the evaluation of our model is predominantly centered on the explicit intentions contained within the provided instructions. Several crucial factors, such as truthfulness and harmlessness \citep{askell2021general}, were not considered in this study. These elements, although not addressed in our current work, represent promising avenues for subsequent research endeavors.

\bibliography{custom}
\bibliographystyle{acl_natbib}

\appendix
\section{Appendix}
\label{sec: appendix}
\subsection{Instruction Dataset Statistics}
\label{sec: appendix_instruction_dataset_statistics}
\begin{figure*}[htbp]
    \subfloat[ ]{{\includegraphics[width=0.45\linewidth]{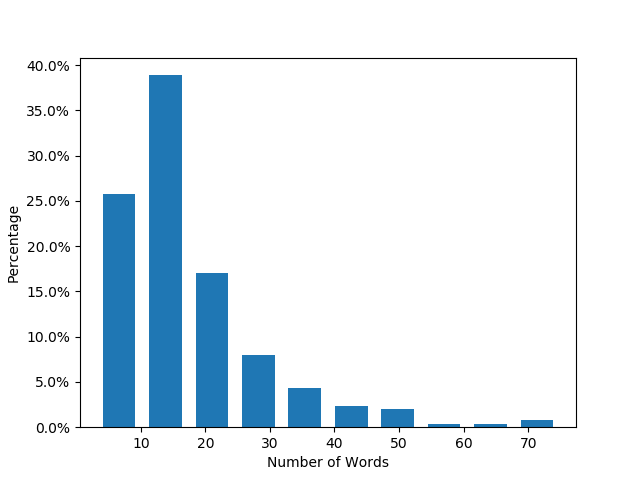} }}
    \hspace{15pt}
    \subfloat[\centering]{{\includegraphics[width=0.45\linewidth]{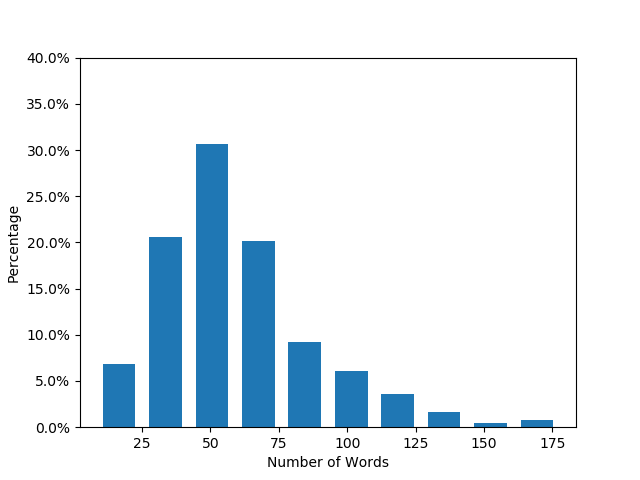} }}%
    \hspace{15pt}
    \subfloat[\centering ]{{\includegraphics[width=0.45\linewidth]{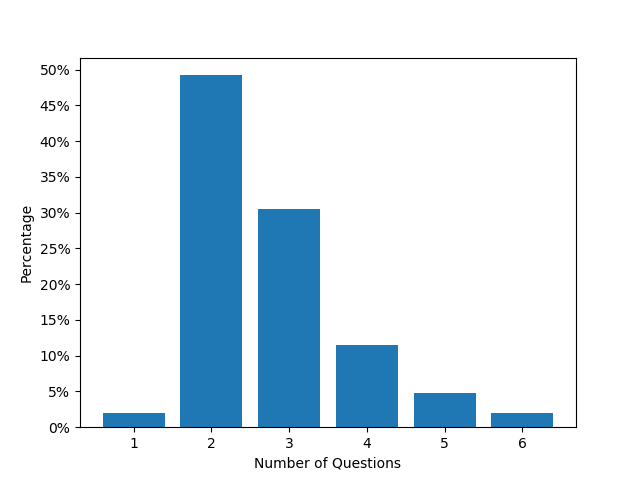} }}%
    \hspace{15pt}
    \subfloat[\centering ]{{\includegraphics[width=0.45\linewidth]{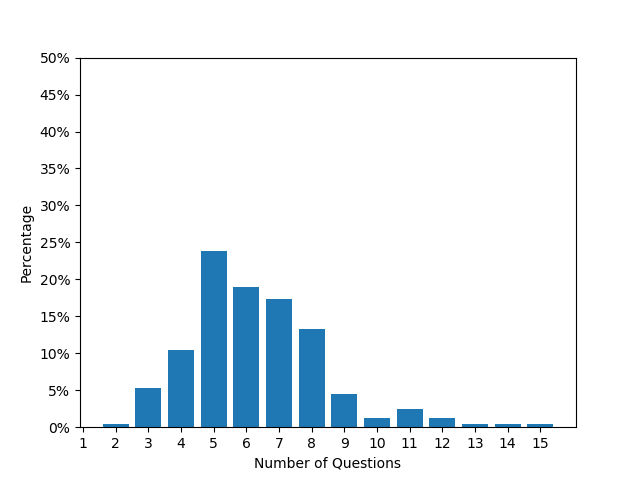} }}%
    \centering
    \caption{Distribution of instruction lengths for the \textit{Easy Set} (a) and \textit{Hard Set} (b) with the x-axis representing the number of words and the y-axis indicating the percentage of instructions. Likewise, the distribution of average questions per instruction for \textit{Easy Set} (c) and \textit{Hard Set} (d) is shown with the x-axis enumerating the number of questions and the y-axis detailing the percentage of instructions.
    }
    \label{fig_instruction_statistics} 
\end{figure*}

Figure~\ref{fig_instruction_statistics} illustrates the distributions pertaining to the instruction length for both the \textit{Easy Set} and \textit{Hard Set}, as well as the average number of questions per instruction for these sets.
The graph indicates that the instruction length for the \textit{Easy Set} predominantly ranges between 10-20 words, while for the \textit{Hard Set} it averages around 50 words.
Similarly, the average number of questions per instruction for the \textit{Easy Set} center is approximately 2-3, whereas, for the \textit{Hard Set}, this number lies within the range of 5-7.
These statistical variations elucidate the enhanced complexity associated with the \textit{Hard Set} as compared to the \textit{Easy Set}.

Table~\ref{tab: appendix_question_label_distribution} presents a quantitative analysis related to Figure~\ref{fig: constrain_type}, depicting the categorization of questions according to their constraint type within the \textit{Easy Set} and \textit{Hard Set}.
This tabulated data underscores the distribution of questions among different constraint types, providing an empirical perspective on the complexity gradation inherent in each set.

\begin{table*}[htbp]
    \centering
    \small
    \begin{tabular}{lrrrrrrr}
    \toprule
    &Content &Format &Number &Style &Linguistic &\#Labels \\
    \midrule
    \textit{Easy Set} &517 &167 &8 &52 &48 &792 \\
    \textit{Hard Set} &878 &627 &411 &138 &120 &2174 \\
    \midrule
    Overall &1395 &794 &419 &190 &168 &2966 \\
    \bottomrule
    \end{tabular}
    \caption{Distribution of Questions Across Different Constraint Types. The '\#Labels' column represents the total count of question labels associated with each dataset. The quantity exceeds the total number of questions listed in Table~\ref{tab: statistics}, as individual questions can potentially be classified under multiple constraint types. }
    \label{tab: appendix_question_label_distribution}
\end{table*}

\subsection{Prompt Design for GPT-4 Automatic Evaluation}
\label{sec: appendix_prompt_design}
In this section, we provide a detailed overview of the prompt design used in our study.
The purpose of these prompts is to guide GPT-4 in evaluating text outputs generated by other LLMs.

Our dataset encompasses two types of instructions: with and without an accompanying input.
For instance, a summarization task might have "Summarize the document" as an "Actual Instruction", and the document content as "Input".
In cases where there is no direct input, the instruction itself guides the LLM, e.g., "Please give me 5 words to describe flowers."
The "Total Instruction" sent to an LLM comprises the concatenation of "Actual Instruction" and "Input" (if present).

The prompt instructs GPT-4 to respond with a simple 'YES' or 'NO' to individually decomposed questions about the LLM's generated text.
This prompt includes the "Input" (if any), the LLM's generated text, and a decomposed question.
Notably, the "Actual Instruction" is not included in the Decomposed Protocol's prompt, as we hypothesize that the decomposed questions encompass the instruction's requirements.
An example is shown in Table~\ref{tab: gpt-4_evaluation_prompt_decomposed}.
As for multiple decomposed questions, We engage in a multi-turn dialogue with GPT-4.
Each instruction and generation is followed by several questions, asked sequentially.
GPT-4's response to each question is used as context for the subsequent question.
We also explored different methods, including asking all questions simultaneously and asking each question independently, but found the sequential multi-turn dialogue to yield more effective results.

We also experimented with incorporating a few-shot learning approach by providing examples in the prompt.
However, the results indicated no significant improvement, possibly due to the varied nature of the instructions and questions, and the simplicity of the answer format.

\begin{table*}[htbp]
    \small
    \centering
    \renewcommand{\arraystretch}{1.5}
    \begin{tabular}{|p{14cm}|}
        \hline
        \textcolor{blue}{\textbf{<User>}:}\\
        Based on the provided Input (if any) and Generated Text, answer the ensuing Questions with either a YES or NO choice. Your selection should be based on your judgment as well as the following rules:\\\\

        - YES: Select 'YES' if the generated text entirely fulfills the condition specified in the question. However, note that even minor inaccuracies exclude the text from receiving a 'YES' rating. As an illustration, consider a question that asks, “Does each sentence in the generated text use a second person?” If even one sentence does not use the second person, the answer should NOT be ‘YES’. To qualify for a 'YES' rating, the generated text must be entirely accurate and relevant to the question.\\\\

        - NO: Opt for 'NO' if the generated text fails to meet the question's requirements or provides no information that could be utilized to answer the question. For instance, if the question asks, “Is the second sentence in the generated text a compound sentence?” and the generated text only has one sentence, it offers no relevant information to answer the question. Consequently, the answer should be ‘NO’.'''\\\\

        Input:\\
        "The typical avocado is over 300 calories from the oil in it. That’s the amount of calories in a large candy bar. If you get enough exercise to eat a large candy bar every day without gaining weight, it wouldn’t be a problem to eat an avocado every day. Other wise you should probably eat them sparingly."\\\\

        Generated Text:\\
        ""Avocados: A Calorie-Dense Treat to Enjoy in Moderation\""\\\\

        Question:\\
        Is the generated text a post title? \\
        
        \textcolor{red}{\textbf{<Assistant>}:}\\
        YES\\
        \textcolor{blue}{\textbf{<User>}:}\\
        Question:\\
        Is the generated text appealing as a post tile? \\
        \textcolor{red}{\textbf{<Assistant>}:}\\
        YES\\
        \textcolor{blue}{\textbf{<User>}:}\\
        Question:\\
        Is the generated post title suitable for the post in the given input?\\
        \textcolor{red}{\textbf{<Assistant>}:}\\
        YES\\

        \hline
    \end{tabular}
    \caption{Example of the multi-turn prompt for automatic evaluation with GPT-4. The "Generated Text" is from Gemini-Pro.}
    \label{tab: gpt-4_evaluation_prompt_decomposed}
\end{table*}

\subsection{Human Annotation Guidance}
\label{sec: appendix_human_annotation_guidance}
\begin{figure*}[htbp]
    \centering
    \includegraphics[width=\linewidth]{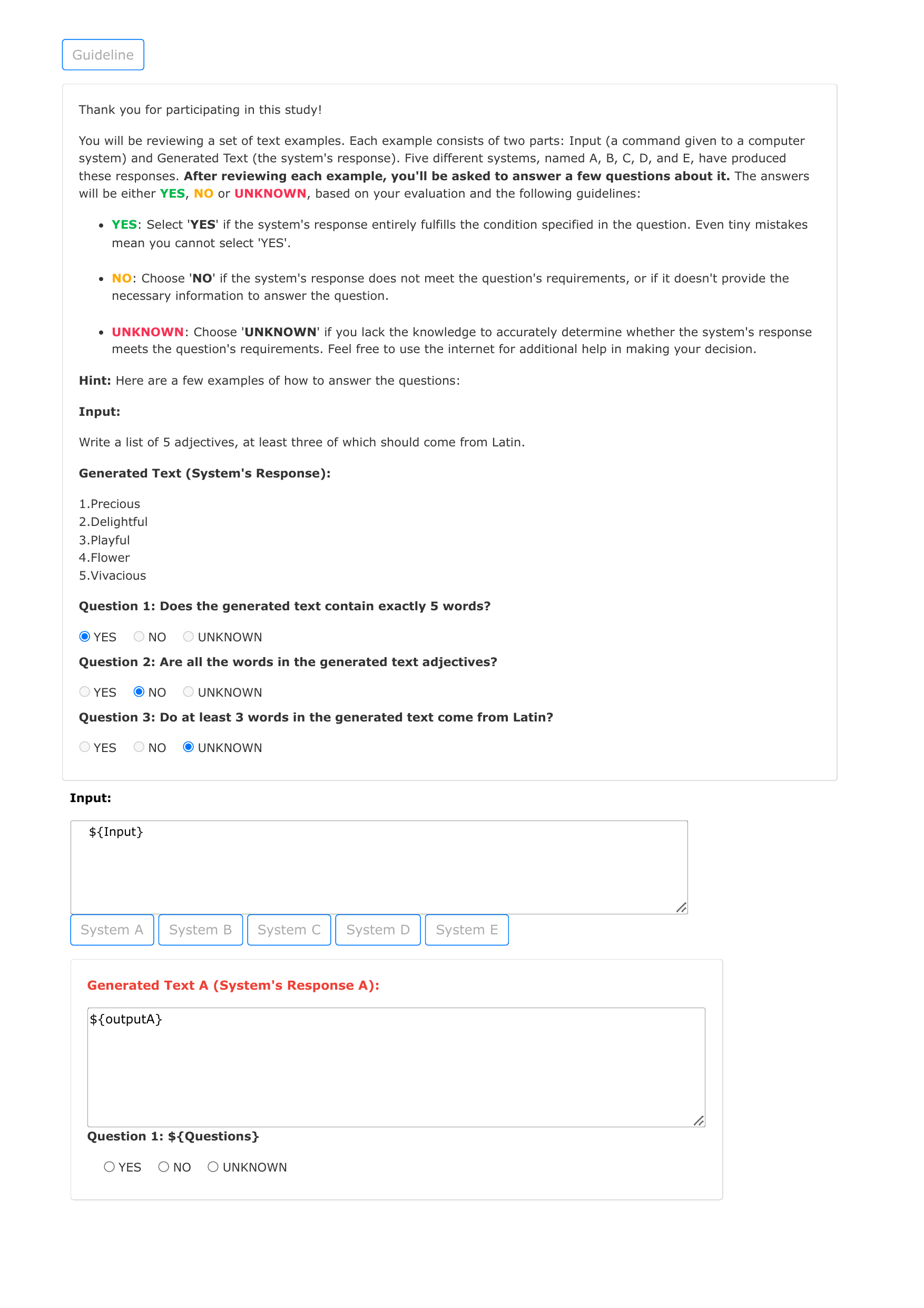}
    \caption{Interface used in Amazon Mechanical Turk}
    \label{fig: AMT interface}
\end{figure*}
The guideline for crowd-sourced workers and interface used in the Amazon Mechanical Turk is demonstrated in Figure~\ref{fig: AMT interface}.
Annotators are asked to read the guidelines before starting the annotation.
We provide one extra option "\texttt{UNKNOWN}" for annotators if they find it hard enough to answer the questions with their knowledge.
Such options will not be counted during the experiment and the evaluation.
Systems are randomly shuffled in each instance to prevent the potential of memorizing which system is better.

\subsection{GPT-4-1106 Evaluation Results}
\label{sec: appendix_GPT-4-1106-preview }
\begin{table}[htbp]
    \centering
    \small
    \setlength\tabcolsep{5.8pt}
    \begin{tabular}{l|ccc}
    \toprule
    Model & \textit{Easy Set} & \textit{Hard Set} & Overall \\
    \midrule
    gpt-4-1106-preview & 85.7 & 85.4 & 85.5 \\
    gpt-3.5-turbo-1106 & 82.5 & 81.2 & 81.6 \\
    claude-2.1 & 77.5 & 81.9 & 80.5 \\
    gemini-pro & 79.7 & 80.2 & 80.1 \\
    Llama-2-70b-chat & 84.5 & 77.6 & 79.7 \\
    Vicuna-13b-v1.5 & 79.1 & 69.5 & 72.4 \\
    \bottomrule
    \end{tabular}
    \caption{Automated \metricname powered by gpt-4-1106-preview for six selected LLMs on the \datasetname.}
    \label{tab: gpt4 evaluation result 1106}
\end{table}

In Table~\ref{tab: gpt4 evaluation result 1106}, we expand our evaluation to incorporate the latest iteration of the GPT-4 model (gpt-4-1106-preview). The findings largely mirror the trends observed in Section~\ref{sec: automatic evaluation}. Notably, the scores attributed by gpt-4-1106-preview are marginally lower than those given by its predecessor, gpt-4-0314. This discrepancy warrants further investigation to understand the underlying factors influencing this variation in scoring.

\begin{figure}[htbp]
\centering
\includegraphics[width=\linewidth]{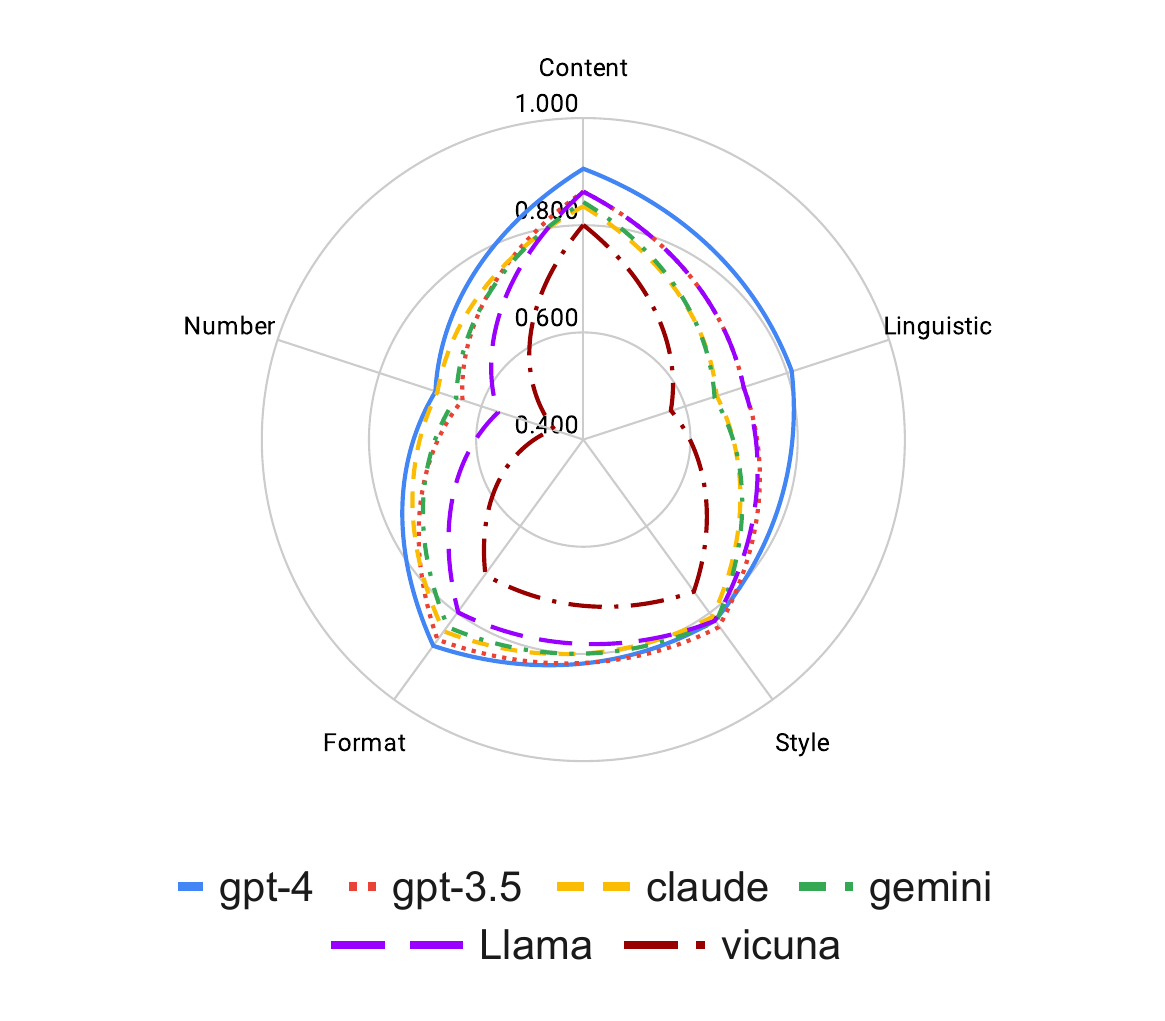}
\caption{Radar plot illustrating the comparative efficacy of various Large Language Models (LLMs) across different constraint types, evaluated using gpt-4-1106-preview.}
\label{fig: constraint 1106}
\end{figure}
\begin{figure}[htbp]
\centering
\includegraphics[width=\linewidth]{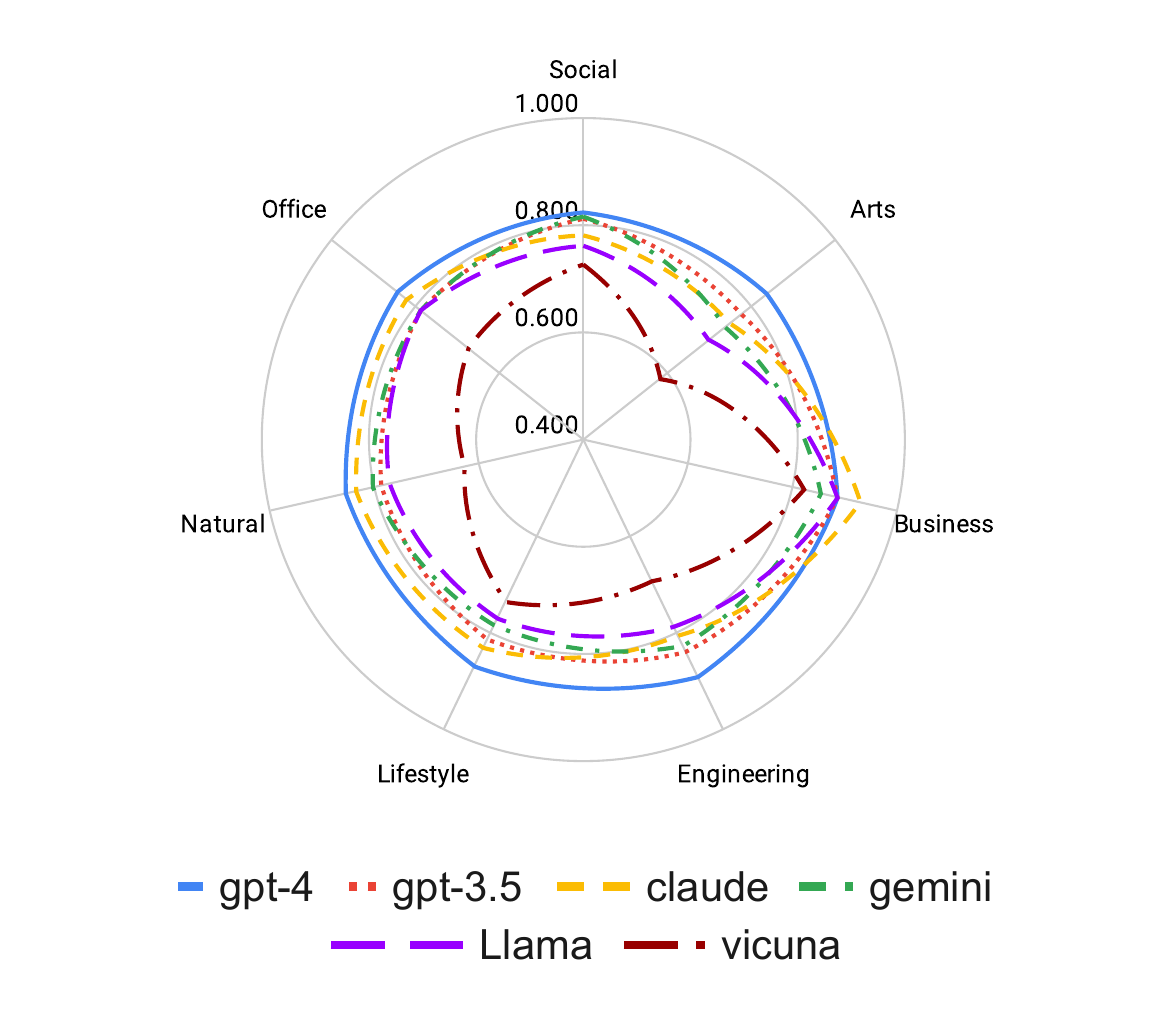}
\caption{Radar plot demonstrating the performance of LLMs in diverse domains, as assessed by gpt-4-1106-preview.}
\label{fig: domain 1106}
\end{figure}

Figures~\ref{fig: constraint 1106} and \ref{fig: domain 1106} present a detailed breakdown of the evaluation results, categorizing performance according to various constraint types and domains. These results are consistent with the findings discussed in Section~\ref{sec: automatic evaluation}.

The comprehensive analysis conducted with the gpt-4-1106-preview model reinforces our understanding of the capabilities and limitations of current LLMs.
Moreover, the consistency in performance trends across different model versions underscores the reliability of GPT-4 as a valuable tool in annotation tasks.

\subsection{GPT-4 Annotation Error Analysis}
\label{sec: appendix_analysis}
This section examines the discrepancies between GPT-4's annotations and those made by humans. We hypothesize two primary reasons for the differences: 1) Inherent ambiguities or subjectivities in the questions that challenge human consensus (e.g., "a letter around 250 words?" where "around" is vague, or "generate an appealing title" where "appealing" is subjective); 2) Actual errors in GPT-4's annotations.

To explore the difficulty in achieving consensus on certain questions, we introduce the concept of a "Disagreement Level" which quantifies the discrepancy between single expert annotation and the gold standard, defined as the majority vote, for each question. This metric serves to illustrate the degree to which human annotators differ in their assessments for each question. Specifically, for a given question and a single LLM's output, we incorporate the binary `YES' or `NO' evaluations from three independent annotators, from which the majority vote is designated as the gold standard. Consequently, for a single LLM's output, the Disagreement Level may be as low as 0 (indicating consensus among all three annotators) or as high as 1 (signifying that one annotator diverged from the decision of the other two). Given that our experimental framework deploys five LMs for each question, the Disagreement Level possesses a potential range of 0 to 5.

\begin{figure}[t]
\centering
\includegraphics[width=7.5cm]{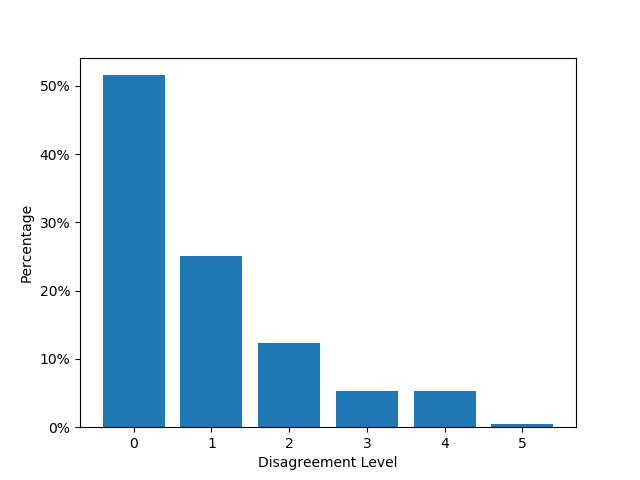}
\caption{Distribution of disagreement levels among annotators. The x-axis represents the level of disagreement among three annotators, while the y-axis indicates the distribution of questions corresponding to each disagreement level.} 
\label{fig: disagreement_level_distribution}
\end{figure}

The distribution of Disagreement Levels, as shown in Figure \ref{fig: disagreement_level_distribution}, reveals that around 30\% to 40\% of questions exhibit a medium to high degree of human disagreement (Disagreement Levels at least 2). In assessing instances where GPT-4's annotations deviates from human consensus, we disregard these questions. This approach stems from the rationale that a lack of human consensus undermines the legitimacy of labeling GPT-4's differing response as erroneous.

We then focus on the remaining 60\% to 70\% of the questions that manifest high agreement among human evaluators, yet GPT-4's annotations do not align. In these cases, we can more confidently identify GPT-4's annotations as errors. Our investigation uncovered several recurring reasons for GPT-4's missteps:

1) The LLMs generations incorporate misleading content. For example:
\begin{example}
Instruction: Generate an eviction notice for a tenant named John Smith, residing at 123 Main Street, Cityville, NY, 87562, for nonpayment of rent for the past two months. The notice should give the tenant 30 days to vacate the property.

Question: Does the generated eviction notice include the tenant's name: John Smith? 

Generation: Generated an eviction notice for tenant John Smith, residing at 123 Main Street, Cityville, NY, 87562, for nonpayment of rent for the past two months. The notice should give the tenant 30 days to vacate the property.

Correct Answer: NO.

GPT-4 Answer: YES.
\end{example}
In this case, the LLM's generation simply reiterates the instruction and does not produce an actual eviction notice. Despite this, the inclusion of the phrase `tenant's name: John Smith' can mislead GPT-4 into assuming that it has correctly completed the task.

2) Requirements necessitating numerical computation or counting. Consider the following instances:

\begin{example}
Instruction: Generate a random alphanumeric string with exactly 20 characters.

Question: Does the generated string exactly include 20 characters? 

Generation: Here is a 20 character random alphanumeric string:

a3k92j7h6r5t1c4b8

Correct Answer: NO.

GPT-4 Answer: YES.

\end{example}
In the illustrated case, GPT-4 is unsuccessful in accurately counting a sequence of 20 characters.

\begin{example}
Instruction: Describe a landscape painting from three distinct layers: foreground, middle ground, and background. Include at least two types of trees, a body of water, and a mountain range. Use a limited color palette of no more than four colors to create harmony, and specify which natural elements correspond to each color. Additionally, provide a description of the textures and brushstrokes used to represent the different natural elements.

Question: Does the painting described by the generated text include at least two types of trees, a body of water, and a mountain range?

Generation: A landscape painting with a foreground of trees and shrubs, a middle ground of rolling hills and a lake, and a background of mountains and trees. The colors used are blues, greens, and oranges to represent the different elements. Soft brushstrokes are used to create a realistic texture of the landscape, while a more bold brushstroke is used to create the mountains.

Correct Answer: NO.

GPT-4 Answer: YES.

\end{example}
Similarly, when tasked with identifying and enumerating two distinct types of trees, GPT-4 may incorrectly categorize repeated references to trees as unique types.

3) The prerequisite for a specific requirement is not met. For instance, 

\begin{example}
Instruction: Please provide five unique first date night ideas, with each catering to different interests and budgets. Include an estimated cost for each idea, along with a brief description of the activity and why it might be appealing to different types of people.

Question: Does each idea in the generated text stand out as unique, each one catering to different budget ranges?

Generation: 
1. A picnic in the park - This is a classic first date idea that is sure to impress. It's budget-friendly and can be customized to fit your interests, whether you want to pack a basket with gourmet snacks or bring a Frisbee to play catch. The cost of a picnic in the park can vary depending on what you bring, but it's generally a very affordable option.

2. Art museum visit - If you and your date are into art, a visit to a local museum can be a great first date idea. Many museums have free admission days, so it won't break the bank. This activity can lead to interesting conversations and can be a great way to get to know each other better.

3. Hiking or biking - If you and your date are outdoorsy, a hike or bike ride can be a fun and unique first date idea. Depending on the location, this activity can range from moderate to intense, so it's important to choose a route that is suitable for both of you. The cost of this activity will depend on the equipment you need to rent, if any.

4. Cooking class - If you and your date love to cook or want to learn, a cooking class can be a great first date idea. Many cooking classes are held in the evening, so it can be a fun and unique way to spend an evening together. The cost of a cooking class can vary depending on the location and the class, but it's generally a more expensive option.

5. Board game cafe - If you and your date are looking for a low-key, indoor activity, a visit to a board game cafe can be a fun and unique first date idea. Many board game cafes have a wide selection of games to choose from, so you can find something that suits your interests. The cost of this activity will depend on how much you spend on food and drinks, but it's generally a more affordable option than a traditional dinner date.

Correct Answer: NO.

GPT-4 Answer: YES.

\end{example}

In this illustration, the generated output fails to satisfy the requirement set forth in the instructions to `include an estimated cost for each idea'. As a result, the generated output does not provide sufficient information to address the question with the requirement `each one catering to different budget ranges'.

4) Requisite Spatial Reasoning Constraints. For illustration, consider the following:
\begin{example}
Instruction: The objective of the Item Elimination Game is to eliminate items in a 6x6 grid by aligning 3 or more of the same item consecutively in a row or column.

Here are the rules:

1. The grid consists of 6 rows and 6 columns, with each cell containing one of three different items represented by A, B, or C.

2. In each step, you may remove items in several cells, provided that they form a consecutive sequence of 3 or more of the same item in the same row or column.

3. Once the item in a cell is removed, it leaves a blank cell, represented by "\#".

Based on the rules, you should generate a 6x6 grid filled with 3 different kinds of items (A, B, and C) as the initial setup and play the game for one step: If you can identify items that form a sequence of 3 or more of the same item in a row or column, leaving blank cells ("\#") in their place and generating the resulting grid; if you cannot find such a sequence, explain it.

Question: Does the generated text include a 6x6 grid as the initial step?

Generation: Here is the initial 6x6 grid filled with 3 different kinds of items (A, B, and C):

```diff

A B C

A B C

A B C

A B C

A B C

A B C

```

After one step of the game, the grid should look like this:

```diff

A B C

A B C

A B C

A B C

A B C

A B C

\#

```

The sequence of 3 or more of the same item that was eliminated is not possible in this case, as there are no consecutive sequences of 3 or more of the same item in the same row or column.

Correct Answer: NO.

GPT-4 Answer: YES.
\end{example}

Within this exemplar, the instruction necessitates the generation of a `6x6 grid'. The generation proclaims `Here is the initial 6x6 grid', yet the rendered grid does not correspond to the stipulated 6x6 dimensions. This instance thus necessitates spatial reasoning proficiency to accurately evaluate whether the produced grid adheres to the 6x6 specification, rather than relying solely on textual interpretation. Being a language model, GPT-4 might demonstrate limitations in spatial reasoning capabilities.

5) The requisite condition remains partially unfulfilled. For instance, consider the following example:
\begin{example}
Instruction: 
Design a 7-day challenge for individuals looking to explore a new hobby: painting. Each day should include a specific task and a difficulty level. The activities should progress in difficulty. Include a brief description (20-30 words) for each day's task.

Question: 
Does every single day of the challenge in the generated text incorporate a particular task along with a corresponding level of difficulty?

Generation: 
Day 1: Start with a blank canvas and a painting medium of your choice. Begin with a simple painting task, such as tracing a design or painting a color wheel. 

Day 2: Paint a landscape or other outdoor scene using light brushstrokes. 

Day 3: Paint a portrait or figure using a photo as reference. 

Day 4: Experiment with abstract art by painting with different materials or using mixed media. 

Day 5: Create a pattern or design by painting repeating shapes or lines. 

Day 6: Paint a still life or indoor scene using shadow and light. 

Day 7: Create a composition using multiple colors and textures.

Correct Answer: NO.

GPT-4 Answer: YES.

\end{example}

In the given scenario, the question necessitates a response containing `a specific task along with a corresponding degree of difficulty'. The generated output successfully encapsulates `a specific task', albeit fails to adhere to the secondary part of the requirement, that is to incorporate `a corresponding degree of difficulty'.

\subsection{Case Study}
As highlighted in section \ref{sec: automatic evaluation}, all models exhibit their lowest performance on Number and Linguistic constraints. Consequently, we selected two specific instructions embodying these constraints for a detailed case study. This section presents the instructions, the related decomposed questions, the responses generated by six chosen models, and the annotation results from GPT-4-0314, GPT-4-1106, as well as expert assessments for the Yes/No responses to all decomposed questions. This analysis aims to provide an intuitive understanding of how each model struggles with the given instructions. For an exhaustive dataset and the complete outputs for all instructions from the six models, please consult our GitHub repository.

Tables \ref{tab: gpt-4_generation_example1_1} to \ref{tab: gpt-4_generation_example1_3} focus primarily on the number constraint. These tables reveal that none of the models fully satisfied the 4th and 5th constraints regarding numbers. This underperformance in the Number category highlights the inherent challenges in numerical reasoning across all LLMs. Specifically, generating a DNA sequence while maintaining the proportion of purines and pyrimidines is complex. Moreover, ensuring equal representation of the four nucleotide types while maintaining this proportion is particularly challenging due to their somewhat contradictory nature.

Tables \ref{tab: gpt-4_generation_example2_1} to \ref{tab: gpt-4_generation_example2_2} are dedicated to the linguistic constraint. In this case, none of the models completely satisfied the 3rd and 4th Linguistic constraints. Notably, GPT-4-1106 demonstrated a commendable attempt to adhere to the "begin with the last letter of the previous word" rule, successfully applying it to most words, albeit compromising the clarity of the plot description (yet still aligning broadly with the Star Wars theme). In contrast, the other models did not seem to attempt this complex task. The suboptimal results in the Linguistic category likely stem from the unique design of our constraints. While LLMs excel in generating grammatically correct and coherent text, they were not optimized for the atypical linguistic patterns introduced by our constraints, like requiring each word to begin with the last letter of the preceding word in this example. Given that most LLMs are trained to predict subsequent words or characters based on conventional linguistic patterns, they struggle when asked to produce content that deviates markedly from these norms.

\begin{table*}[!htp]\centering
\renewcommand{\arraystretch}{1.5}  
\small
\begin{tabular}{|p{1.5cm}|p{7cm}|p{1.5cm}|p{1.5cm}|p{1.5cm}|}
\hline
\multicolumn{5}{|p{15cm}|}{ID: domain\_oriented\_task\_31; Category: Natural Sciences: Biology} \\
\hline
\multicolumn{5}{|p{15cm}|}{Instruction: Generate a double-stranded DNA consisting of 24 nucleotides in each strand, with an equal number of adenine (A), thymine (T), cytosine (C), and guanine (G) nucleotides. For every 6 nucleotides in the first strand, the first 3 should have a higher proportion of purines (A and G), while the last 3 should have a higher proportion of pyrimidines (C and T). The second strand should be complementary to the first strand, following the base pairing rules (A pairs with T, and C pairs with G).} \\
\hline

\multicolumn{5}{|p{15cm}|}{Decomposed Questions:\newline
1. Is the generated sequence a double-stranded DNA? (Format) \newline
2. Does each strand of the generated double-stranded DNA consist of 24 nucleotides? (Number) \newline
3. In the generated double-stranded DNA, is there an equal number of adenine (A), thymine (T), cytosine (C), and guanine (G) nucleotides? (Number) \newline
4. For every set of 6 nucleotides in the first strand of the generated double-stranded DNA, do the first 3 have a higher proportion of purines (A and G)? (Number) \newline
5. For every set of 6 nucleotides in the first strand of the generated double-stranded DNA, do the last 3 have a higher proportion of pyrimidines (C and T)? (Number) \newline
6. Is the second strand of the generated double-stranded DNA complementary to the first strand, adhering to the base pairing rules where adenine (A) pairs with thymine (T), and cytosine (C) pairs with guanine (G)? (Format)  } \\
\hline

Model-Name & Model-Generation & GPT-4-0314 Annotation & GPT-4-1106-Annotation & Expert-Annotation \\
\hline
GPT-4-1106 & To create a double-stranded DNA with the specified conditions, we need to ensure that the first strand has an equal number of each nucleotide (A, T, C, G) and that every 6-nucleotide segment starts with a purine-rich triplet followed by a pyrimidine-rich triplet. The second strand will be complementary to the first.\newline\newline

Here's an example of such a DNA sequence:\newline\newline

First strand (5' to 3'):\newline
AGCCTAGTCGACTAGCTAGCCGAT\newline\newline

Second strand (3' to 5', complementary to the first strand):
TCGGATCAGCTGATCGATCGGCTA\newline\newline

Breaking down the first strand into 6-nucleotide segments, we have:\newline
- AGCCTA (purine-rich: AG, pyrimidine-rich: CTA)\newline
- GTCGAC (purine-rich: GT, pyrimidine-rich: CGA)\newline
- TAGCTA (purine-rich: TA, pyrimidine-rich: GCT)\newline
- GCCGAT (purine-rich: GC, pyrimidine-rich: GAT)\newline\newline

Each segment starts with a higher proportion of purines (A and G) and ends with a higher proportion of pyrimidines (C and T). The second strand is the perfect complement to the first, following the base pairing rules (A with T and C with G). & [Yes, Yes, No, Yes, Yes, Yes] & [Yes, Yes, No, Yes, No, Yes] & [Yes, Yes, No, No, No, Yes] \\\hline
gpt-3.5-turbo-1106 & First strand:\newline
5'- AAGCTTCCGGAATTCCGGAAGCTT -3'\newline\newline

Second strand:\newline
3'- TTCGAAGGCCTTAAGGCCTTCGAA -5' & [Yes, Yes, No, Yes, No, Yes] & [Yes, Yes, False, No, No, Yes] & [Yes, Yes, Yes, No, No, Yes] \\
\bottomrule
\end{tabular}
\caption{Model Generation Example 1.}\label{tab: gpt-4_generation_example1_1}
\end{table*}

\begin{table*}[!htp]\centering
\renewcommand{\arraystretch}{1.5}  
\small
\begin{tabular}{|p{1.5cm}|p{7cm}|p{1.5cm}|p{1.5cm}|p{1.5cm}|}
\hline
claude-2.1&  Here is a randomly generated 24 nucleotide double-stranded DNA sequence with the requested properties:\newline\newline

Strand 1: AGGCTA CTTAGC GGAATC TTCGCA\newline
Strand 2: TCCGAT GAATCG CCTTAG AAGCGT &[Yes, Yes, No, Yes, Yes, Yes]&[Yes, Yes, No, Yes, No, Yes]&[Yes, Yes, Yes, No, No, Yes]\\\hline
gemini-pro&"Strand 1 (5' to 3'):\newline\newline

**Purine-rich region:**\newline
- AGTACG\newline\newline

**Pyrimidine-rich region:**\newline
- TCGAAT\newline\newline

**Purine-rich region:**\newline
- AGCTAC\newline\newline

**Pyrimidine-rich region:**\newline
- TCGATT\newline\newline

Strand 2 (3' to 5'):\newline

**Pyrimidine-rich region:**\newline
- TCGAAT\newline\newline

**Purine-rich region:**\newline
- AGCTAC\newline\newline

**Pyrimidine-rich region:**\newline
- TCGATT\newline\newline

**Purine-rich region:**\newline
- AGTACG"&[Yes, No, No, Yes, Yes, No]&[Yes, No, No, Yes, No, No]&[Yes, Yes, No, No, No, No]\\
\bottomrule
\end{tabular}
\caption{Model Generation Example 1. (continued)}\label{tab: gpt-4_generation_example1_2}
\end{table*}

\begin{table*}[!htp]\centering
\renewcommand{\arraystretch}{1.5}  
\small
\begin{tabular}{|p{1.5cm}|p{7cm}|p{1.5cm}|p{1.5cm}|p{1.5cm}|}
\hline
vicuna-13b-v1.5 &Here is a possible double-stranded DNA sequence that meets the given criteria:

First strand:\newline
ATCGGATCGATCGATCGATCGATCGATCGATCGAT\newline
CGATCGATCGATCGATCGATCGATCGATCGATCGA\newline
TCGATCGATCGATCGATCGATCGATCGATCGATCG\newline
ATCGATCGATCGATCGATCGATCGATCGATCGATC\newline
GATCGATCGATCGATCGATCGATCGATCGATCGAT\newline
CGATCGATCGATCGATCGATCGATCGATCGATCGA\newline
TCGATCGATCGATCGATCGATCGATCGATCGATCG\newline
ATCGATCGATCGATCGATCGATCGATCGATCGATC\newline
GATCGATCGATCGATCGATCGATCGATCGATCGAT\newline
CGATCGATCGATCGATCGATCGATCGATCGATCGA\newline
TCGATCGATCGATCGATCGATCGATCGATCGATCG\newline
ATCGATCGATCGATCGATCGATCGATCGATCGATC\newline
GATCGATCGATCGATCGATCGATCGATCGATCGAT\newline
CGATCGATCGATCGATCGATCGATCGATCGATCGA\newline
TCGATCGATCGATCGATCGATCGATCGATCGATCG\newline
ATCGATCGATCGATCGATCGATCGATCGATCGATC\newline
GATCGATCGATCGATCGATCGATCGATCGATCGAT\newline
CGATCGATCGATCGATCGATCGATCGATCGATCGA\newline
TCGATCGATCGATCGATCGATCGATCGATCGATCG\newline
ATCGATCGATCGATCGATCGATCGATCGATCGATC\newline
GATCGATCGATCGATCGATCGATCGATCGATCGAT\newline
CGATCGATCGATCGATCGATCGATCGATCGATCGA\newline
TCGATCGATCGATCGATCGATCGATCGATCGATCG\newline
ATCGATCGATCGATCGATCGATCGATCGATCGATC\newline
GATCGATCGATCGATCGATCGATCGATCGATCGAT\newline
CGATCGATCGATCGATCGATCGATCGATCGATCGA\newline
TCGATCGATCGATCGATCGATCGATCGATCGATCG\newline
......(omit some repeated 'ATCG's) &[Yes, No, No, Yes, Yes, No]&[Yes, No, No, No, No, No]&[No, No, No, No, No, No]\\\hline
Llama-2-70b-chat-hf & Sure, here's a possible double-stranded DNA molecule with 24 nucleotides in each strand, having an equal number of A, T, C, and G nucleotides, and following the purine-pyrimidine pattern you described:\newline\newline

Strand 1:\newline\newline

ATCGGCTCTGCAGAGTCTCTG\newline\newline

Strand 2:\newline\newline

TGCAGGCTCTAGCTCTGCAGT\newline\newline

In Strand 1, the first 3 nucleotides (ATC) have a higher proportion of purines, while the last 3 nucleotides (GCT) have a higher proportion of pyrimidines. This pattern is repeated every 6 nucleotides.\newline\newline

Strand 2 is complementary to Strand 1, following the base pairing rules. The first 3 nucleotides (TGC) pair with the last 3 nucleotides of Strand 1 (GCT), while the last 3 nucleotides (AGT) pair with the first 3 nucleotides of Strand 1 (ATC).\newline\newline

I hope this helps! Let me know if you have any further questions. & [Yes, No, No, No, No, No]&[Yes, No, No, No, No, No]&[Yes, No, No, No, No, No]\\
\bottomrule
\end{tabular}
\caption{Model Generation Example 1. (continued)}\label{tab: gpt-4_generation_example1_3}
\end{table*}

\begin{table*}[!htp]\centering
\renewcommand{\arraystretch}{1.5}  
\small
\begin{tabular}{|p{1.5cm}|p{7cm}|p{1.5cm}|p{1.5cm}|p{1.5cm}|}
\hline
\multicolumn{5}{|p{15cm}|}{ID: domain\_oriented\_task\_0; Category: Arts: Film} \\
\hline
\multicolumn{5}{|p{15cm}|}{Instruction: Generate a sentence that describes the plot of Star Wars, where each word begins with the end letter of the previous word, and the set of beginning letters of all words in the sentence covers every letter in the alphabet from A to Z.} \\
\hline

\multicolumn{5}{|p{15cm}|}{Decomposed Questions:\newline
1. Is the generated text a sentence? (Format, Number)\newline
2. Does the generated text describe the plot of Star Wars? (Content) \newline
3. In the generated sentence, does each word begin with the last letter of the pre- vious word? (Linguistic) \newline
4. In the generated sentence, do the be- ginning letters of all words collectively cover every letter in the English alphabet from A to Z? (Linguistic)} \\
\hline

Model-Name & Model-Generation & GPT-4-0314 Annotation & GPT-4-1106-Annotation & Expert-Annotation \\
\hline
GPT-4-1106 & Anakin's saga advances, securing galactic control, leaving legions stormtrooping, gaining ground, destroying galaxies, securing Sith's supremacy, yet triumphing, gaining new wisdom, mastering galactic conflicts, saving generations, securing eternal equilibrium. & [Yes, Yes, Yes, No] & [No, No, Yes, No] & [Yes, Yes, No, No] \\\hline
gpt-3.5-turbo-1106 & A long time ago, in a galaxy far, far away, Yoda and Anakin navigated through the stars, searching for Sith, hoping to overthrow the evil empire, eventually leading to the destruction of the Death Star, bringing peace to the galaxy. & [Yes, Yes, No, No] & [Yes, Yes, No, No] & [Yes, Yes, No, No] \\\hline
claude-2.1 &  Here is an attempt at a 26-word sentence for the plot of Star Wars with each word chaining on the previous word's last letter:\newline\newline

A long time ago in a galaxy far, far away, rebels battle an evil empire's death star space station. Quietly, a young man named Luke trains to become a Jedi knight under Obi-Wan's guidance. X-wings attack the station, which explodes spectacularly. Luke and friends triumph gloriously, though Darth Vader escapes narrowly. Victory comes at a cost for the alliance bravely fighting imperial oppression zealously. &[No, Yes, No, No]&[No, Yes, No, No]&[No, Yes, No, No]\\\hline
gemini-pro &Zealous Jedi Knights battle menacing Sith Lords, causing galactic unrest.&[Yes, Yes, No, No]&[Yes, Yes, No, No]
&[Yes, Yes, No, No]\\\hline
vicuna-13b-v1.5&A B C D E F G H I J K L M N O P Q R S T U V W X Y Z: A young farm boy, Luke Skywalker, joins forces with a rogue smuggler, Han Solo, and a strong-willed princess, Leia Organa, to destroy a powerful space station, the Death Star, and save the galaxy from the evil Empire's clutches, ultimately leading to a climactic battle between good and evil.&[Yes, Yes, No, No]&[Yes, Yes, No, No]&[Yes, Yes, No, No]\\
\bottomrule
\end{tabular}
\caption{Model Generation Example 2.}\label{tab: gpt-4_generation_example2_1}
\end{table*}

\begin{table*}[!htp]\centering
\renewcommand{\arraystretch}{1.5}  
\small
\begin{tabular}{|p{1.5cm}|p{7cm}|p{1.5cm}|p{1.5cm}|p{1.5cm}|}
\hline
Llama-2-70b-chat-hf&Sure! Here's a sentence that describes the plot of Star Wars using your criteria:\newline\newline

A long time ago, in a galaxy far, far away, a brave Alliance pilot named Luke Skywalker, skilled in the ways of the Force, embarked on a daring mission to rescue Princess Leia from the evil Empire's dreaded Death Star, fighting off fierce stormtroopers and dodging laser blasts along the way.\newline\newline

Here's how the sentence covers every letter of the alphabet from A to Z:\newline\newline

A - A\newline
B - Brave\newline
C - Cunning\newline
D - Daring\newline
E - Empire's\newline
F - Fierce\newline
G - Galaxy\newline
H - Heroic\newline
I - In\newline
J - Jedi\newline
K - Killer\newline
L - Laser\newline
M - Mission\newline
N - Nimble\newline
O - Offensive\newline
P - Pilot\newline
Q - Quest\newline
R - Rescue\newline
S - Skilled\newline
T - Threatening\newline
U - Universe\newline
V - Valiant\newline
W - Ways\newline
X - X-wing\newline
Y - Yoda\newline
Z - Zealous\newline\newline

I hope that helps!&[No, Yes, No, No]&[No, Yes, No, No]&[Yes, Yes, No, No]\\
\bottomrule
\end{tabular}
\caption{Model Generation Example 2. (continued)}\label{tab: gpt-4_generation_example2_2}
\end{table*}

\end{document}